\documentclass[
11pt, % The default document font size, options: 10pt, 11pt, 12pt
english, % ngerman for German
singlespacing, % Single line spacing, alternatives: onehalfspacing or doublespacing
headsepline, % Uncomment to get a line under the header
]{MastersDoctoralThesis} % The class file specifying the document structure

\usepackage[utf8]{inputenc} % Required for inputting international characters
\usepackage[T1]{fontenc} % Output font encoding for international characters
\usepackage[ruled,lined,linesnumbered]{algorithm2e}
\DontPrintSemicolon

\SetKwComment{tcc}{$\triangleright$~}{}
\SetCommentSty{normalfont}
\SetKwInput{KwRequire}{Require}
\SetNlSty{}{}{:}

\newcommand*{\Comb}[2]{{}^{#1}C_{#2}}%
\usepackage{mathpazo} % Use the Palatino font by default

\usepackage[backend=bibtex,style=authoryear,natbib=true]{biblatex} % Use the bibtex backend with the authoryear citation style (which resembles APA)

\usepackage[autostyle=true]{csquotes} % Required to generate language-dependent quotes in the bibliography

\thesistitle{Role of Morphogenetic Competency on Evolution} % Your thesis title, this is used in the title and abstract, print it elsewhere with \ttitle
\supervisor{Dr. Michael Levin} % Your supervisor's name, this is used in the title page, print it elsewhere with \supname
\examiner{} % Your examiner's name, this is not currently used anywhere in the template, print it elsewhere with \examname
\degree{Masters of Research in Cognitive Science} % Your degree name, this is used in the title page and abstract, print it elsewhere with \degreename
\author{Lakshwin Shreesha M K} % Your name, this is used in the title page and abstract, print it elsewhere with \authorname
\addresses{} % Your address, this is not currently used anywhere in the template, print it elsewhere with \addressname

\subject{Biological Sciences} % Your subject area, this is not currently used anywhere in the template, print it elsewhere with \subjectname
 % Keywords for your thesis, this is not currently used anywhere in the template, print it elsewhere with \keywordnames
\university{\href{https://u-paris.fr/en/}{Université Paris Cité}} % Your university's name and URL, this is used in the title page and abstract, print it elsewhere with \univname
\department{\href{https://cogmaster.ens.psl.eu/en}{CogMaster}} % Your department's name and URL, this is used in the title page and abstract, print it elsewhere with \deptname
\group{\href{https://drmichaellevin.org/}{Levin Lab}} % Your research group's name and URL, this is used in the title page, print it elsewhere with \groupname
\faculty{\href{http://faculty.university.com}{}} % Your faculty's name and URL, this is used in the title page and abstract, print it elsewhere with \facname

\AtBeginDocument{
\hypersetup{pdftitle=\ttitle} % Set the PDF's title to your title
\hypersetup{pdfauthor=\authorname} % Set the PDF's author to your name
\hypersetup{pdfkeywords=\keywordnames} % Set the PDF's keywords to your keywords
}

\begin{document}

\frontmatter % Use roman page numbering style (i, ii, iii, iv...) for the pre-content pages

\pagestyle{plain} % Default to the plain heading style until the thesis style is called for the body content

%----------------------------------------------------------------------------------------
%	TITLE PAGE
%----------------------------------------------------------------------------------------

\begin{titlepage}
\begin{center}

%\vspace*{.06\textheight}
%{\scshape\LARGE \univname\par}\vspace{1.5cm} % University name
%\textsc{\Large  Thesis}\\[0.5cm] % Thesis type

%{\Huge \bfseries \ttitle\par}\vspace{0.4cm} % Thesis title
%\vspace*{.06\textheight}
%{\scshape\LARGE \authorname\par}\vspace{1.5cm} % University name

\HRule \\[0.4cm] % Horizontal line
{\Huge \bfseries \ttitle\par}\vspace{0.4cm} % Thesis title
\HRule \\[1.5cm] % Horizontal line

%\vspace*{.06\textheight}
{\scshape\Large \authorname\par}\vspace{1.5cm} % University name

{\scshape \emph{supervised by:}\par}
{\scshape\large \supname\par}\vspace{0.1cm} % University name
{\scshape \href{https://drmichaellevin.org/}{The Levin Lab}}\vspace{1.5cm}

\includegraphics[width=6.5cm]{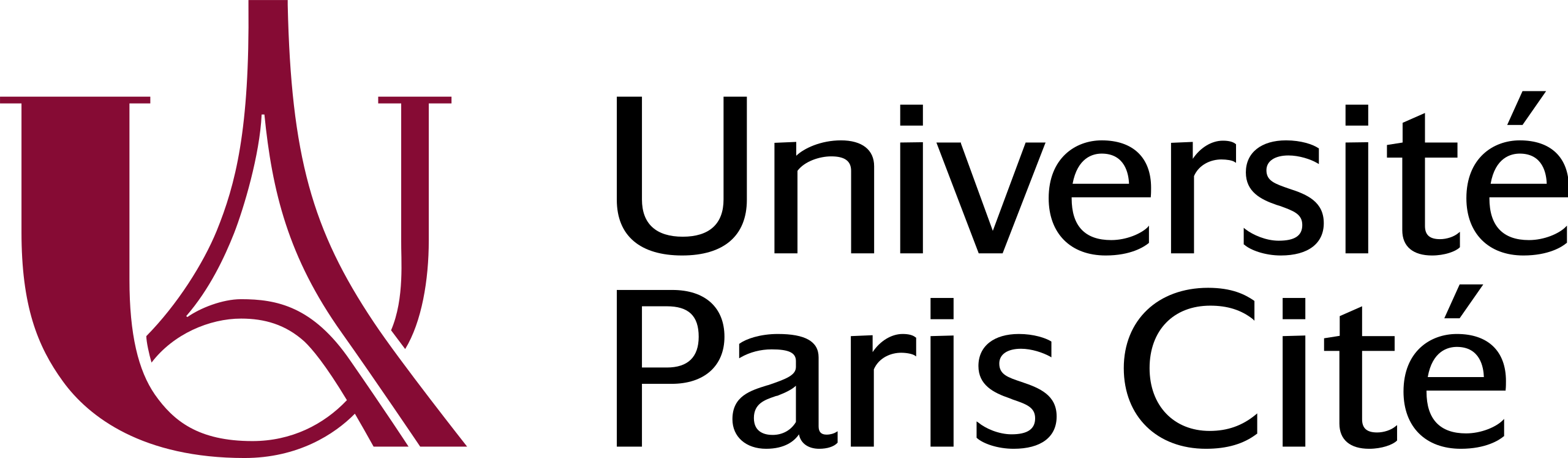}\hspace{1cm}
\includegraphics[width=6.5cm]{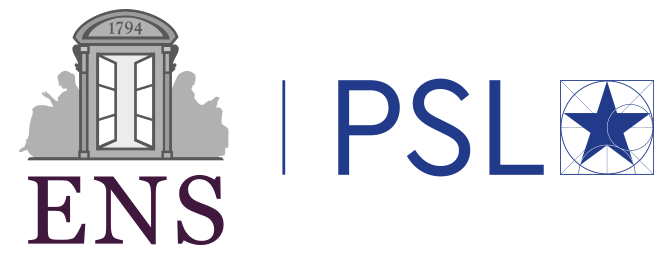}

\vspace{1cm}

\includegraphics[width=2.5cm]{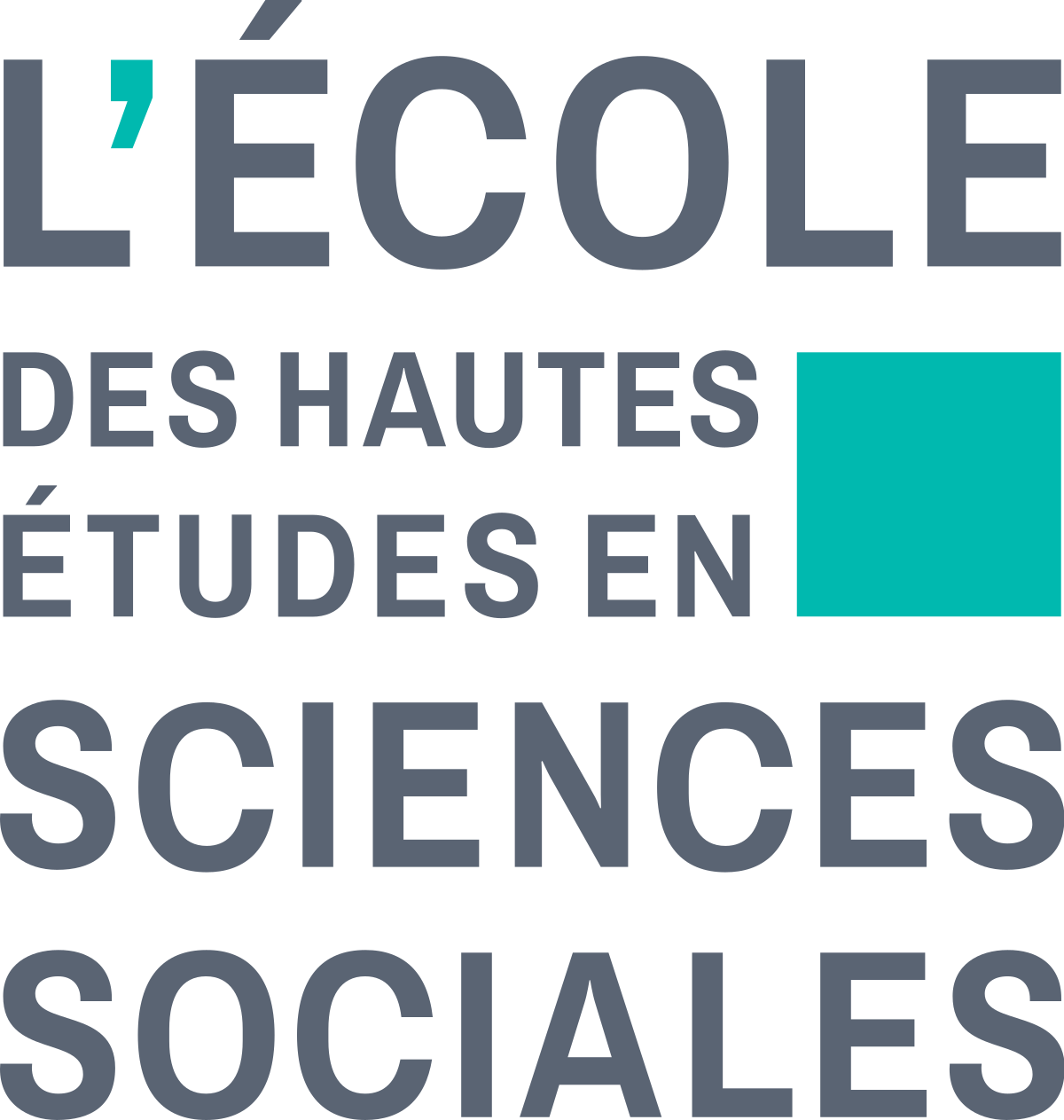}

\vspace{2.5cm}

\Large Thesis submitted for the degree of\\ \degreename\\ 30 Credits\\[0.3cm] % University requirement text
\vspace{0.5cm}
\deptname\\
%\groupname\\\deptname\\[2cm] % Research group name and department name

\vspace{1cm}
 
{\large Spring 2023}\\[4cm] % Date

\vfill
\end{center}
\end{titlepage}

%----------------------------------------------------------------------------------------
%	DECLARATION PAGE
%----------------------------------------------------------------------------------------

\chapter{Declaration of Originality} % Main appendix title

\label{originalitydoc} % For referencing this appendix elsewhere, use \ref{AppendixA}
\setlength\parindent{0pt}

The relationship between intelligence and evolution is bidirectional. The convectional idea of evolution is to treat it as a process which improves the intelligence of a population. Equally important is the fact that the degree of intelligence itself impacts evolution. This inverse relationship is studied in literature from the perspective of behavior i.e., how does lifetime-behavior of an individual effect evolution? We extend this view by studying it from the perspective of morphogenesis i.e., how does morphogenesis impact evolution?\\

Based on observations in Developmental Biology, we believe that morphogenesis is a highly competent process, carried out by cells which behave \textit{not} as passive materials but as active goal-directed agents. Given such morphogenetic competency during evolution, we study how it influences evolutionary dynamics. To this end, we carry out computer simulations using a classic genetic algorithm.\\

Understanding the impact of morphogenetic competency on evolution can help us further our understanding in the fields of Evolutionary Computation and Developmental Biology.

\begin{itemize}
    \item If morphogenesis proves beneficial to evolutionary search, it can be incorporated in genetic algorithms, potentially encouraging the \textit{growth} of problem-solving agents.
    
    \item It can complement approaches which study the inverse relationship from the higher order perspective of behavior.
    
    \item Understanding how evolution regulates morphogenesis could help shed light on confounding morphological phenomena in Biology. For instance, Planaria are a class of flatworms with a messy genome -- they have different chromosomes in each cell. Despite this, each fragment of Planaria regenerates to a complete worm with remarkable accuracy. How does an organism with the most chaotic genome have the best regenerative fidelity, immortality and cancer resistance?
\end{itemize} 
\chapter{Declaration of Contribution} % Main appendix title

\label{contribution} % For referencing this appendix elsewhere, use \ref{AppendixA}

\newcommand{\LS}{\textbf{L.S }}

Primary contributors: Lakshwin Shreesha (\LS) and Dr. Michael Levin (M.L)

\begin{itemize}
    \item Definition of the research question: Initial conceptualization by M.L, refined by \LS with supervision from M.L.

    \item Literature Review: \LS with supervision from M.L

    \item Methodology: Initial idea by M.L, structured into a framework by \LS with supervision from M.L

    \item Software: \LS

    \item Data Analysis: \LS and M.L

    \item Data Curation: \LS

    \item Interpretation of results: \LS and M.L

    \item Thesis writing: \LS 

    \item Thesis proofreading: M.L

    \item Discussions on statistical analyses: Dr. Santosh Manicka, Dr. Hananel Hazan

\end{itemize}

We would like to make it explicit here that Chapters \ref{Chapter3}: Methodology and \ref{Chapter4}: Results,  presented in this thesis borrow heavily from a recently published manuscript \parencite{Shreesha2023} whose primary authors were \LS and M.L. Author contributions for the manuscript remain the same as those mentioned above.\\

What is different here is the perspective from which these results are presented. Chapter \ref{Chapter1}: Introduction, motivates the broader direction in which we intend to head with this line of research, and the literature review (Chapter \ref{Chapter2}) complements it. We discuss implications of our results (Chapter \ref{Chapter5}) from this renewed perspective and conclude (Chapter \ref{Chapter6}) by placing it in context with our future goals.   
 \label{d-contribution}
\cleardoublepage

%----------------------------------------------------------------------------------------
%	ABSTRACT PAGE
%----------------------------------------------------------------------------------------
\chapter{Abstract} % Main appendix title

\label{Abstract} % For referencing this appendix elsewhere, use \ref{AppendixA}

The relationship between intelligence and evolution is bidirectional: while evolution can help evolve intelligences, the degree of intelligence itself can impact evolution \parencite{Baldwin1896}. In the field of Evolutionary Computation, the inverse relationship (impact of intelligence on evolution) is approached from the perspective of organism level behaviour \parencite{Hinton1996}. We extend these ideas to the developmental (cellular morphogenetic) level in the context of an expanded view of intelligence as not only the ability of a system to navigate the three-dimensional world, but also as the ability to navigate other arbitrary spaces (transcriptional, anatomical, physiological, etc.)\parencite{Fields2022}. Here, we specifically focus on the intelligence of a minimal model of a system navigating anatomical morphospace, and assess how the degree and manner of problem solving competency during morphogenesis effects evolutionary dynamics. To this end, we evolve populations of artificial embryos using a standard genetic algorithm in silico. Artificial embryos were cellular collectives given the capacity to undergo morphogenetic rearrangement (e.g., regulative development) prior to selection within an evolutionary cycle. Results from our model indicates that morphogenetic competency  significantly alters evolutionary dynamics, with evolution preferring to improve anatomical intelligence rather than perfect the structural genes. These observations hint that evolution in the natural world may be leveraging the problem solving competencies of cells at multiple scales to boost evolvability and robustness to novel conditions. We discuss implications of our results for the Developmental Biology and Artificial Life communities.

%\include{other-chapters/pre-registration}

%----------------------------------------------------------------------------------------
%	ACKNOWLEDGEMENTS
%----------------------------------------------------------------------------------------

\begin{acknowledgements}
\addchaptertocentry{\textbf{\acknowledgementname}} % Add the acknowledgements to the table of contents
I would like to thank Dr. Michael Levin, my mother, Jyothi TK, and the CogMaster program for their infinite patience in teaching me. It is in their collective wisdom that I strive to make up for my ignorance.\\

Any piece of creative work is more than the sum of its parts: its the instances which have shaped the authors experiences. I would like to thank a dear friend, Jean-Victor Steinlein for significantly having shaped mine over the past two years. Our exchanges alone have made my time at the CogMaster memorable.\\

Finally, I'd like to thank the SMARTS-UP scholarship for their support over the past two years. This thesis would not have been what it is, if it weren't for them.

\end{acknowledgements}

%----------------------------------------------------------------------------------------
%	LIST OF CONTENTS/FIGURES/TABLES PAGES
%----------------------------------------------------------------------------------------

\tableofcontents % Prints the main table of contents

\listoffigures % Prints the list of figures

\listoftables % Prints the list of tables

%----------------------------------------------------------------------------------------
%	ABBREVIATIONS
%----------------------------------------------------------------------------------------

\begin{abbreviations}{ll} % Include a list of abbreviations (a table of two columns)

\textbf{GA} & \textbf{G}enetic \textbf{A}lgorithm\\
\textbf{GAs} & \textbf{G}enetic \textbf{A}lgorithm\textbf{s}\\
\textbf{NN} & Artificial \textbf{N}eural \textbf{N}etwork\\
\textbf{NNs} & Artificial \textbf{N}eural \textbf{N}etwork\textbf{s}\\
\textbf{BP} & \textbf{B}ack \textbf{P}ropagation\\
\textbf{3D} & \textbf{T}hree \textbf{D}imensional\\

\end{abbreviations}

%----------------------------------------------------------------------------------------
%	DEFINITIONS
%----------------------------------------------------------------------------------------

\chapter{Definitions} % Main appendix title

\label{definitions} % For referencing this appendix elsewhere, use \ref{AppendixA}

\begin{itemize}
    \item\textbf{Intelligence}: Problem solving ability by an entity.

    \item \textbf{Morphogenesis}: A process of anatomical change which transforms a disorganized collective into an ordered shape.

    \item \textbf{Evolution}: Darwinian evolution, wherein a population improves its performance in solving a specific task -- as assessed by a fitness function -- over time. 

    \item \textbf{Development}: A process during which morphogenesis is said to take place.

    \item \textbf{Embryo}: A disorganized, jumbled cellular collective.

    \item \textbf{Individual}: An organized structure produced post morphogenesis of an embryo.

    \item \textbf{Hardwired}: A term used to describe the incapability of an embryo to undergo morphogenesis. 

    \item \textbf{Competent}: A term attributed to the ability of an embryo's cells to behave as agential components while undergoing morphogenesis.

    \item  \textbf{Morphogenetic Competency}: If an embryo, composed of goal-directed cells, undergoes morphogenesis, it is said to exhibit morphogenetic competency. 
 
\end{itemize}
%-----------------------------------------------------------------
%----------------------------------------------------------------------------------------
%	THESIS CONTENT - CHAPTERS
%----------------------------------------------------------------------------------------

\mainmatter % Begin numeric (1,2,3...) page numbering

\pagestyle{thesis} % Return the page headers back to the "thesis" style

% Include the chapters of the thesis as separate files from the Chapters folder
% Uncomment the lines as you write the chapters

% Chapter 1

\chapter{Introduction} % Main chapter title
\label{Chapter1} % For referencing the chapter elsewhere, use \ref{Chapter1} 

%----------------------------------------------------------------------------------------

% Define some commands to keep the formatting separated from the content 
\newcommand{\keyword}[1]{\textbf{#1}}
\newcommand{\tabhead}[1]{\textbf{#1}}
\newcommand{\code}[1]{\texttt{#1}}
\newcommand{\file}[1]{\texttt{\bfseries#1}}
\newcommand{\option}[1]{\texttt{\itshape#1}}

%----------------------------------------------------------------------------------------

Evolutionary computation is a field in computer science which adopts techniques from biological evolution to solve computational problems. A common class of algorithms employed in the field are genetic algorithms \parencite{Holland1992}. These algorithms serve to evolve entities with progressively increasing problem solving capabilities (i.e., intelligence). A crucial aspect considered here is the bidirectional relationship between evolution and intelligence: while evolution can help improve intelligence, the degree of intelligence itself impacts evolutionary dynamics \parencite{Baldwin1896, Waddington1953}.\\

A majority of work in literature focuses on the forward relationship (evolution of intelligence), but much less is known about the \textit{inverse relationship} (intelligence’s impacts on evolution). A small body of work does exist which acknowledges the inverse relationship, but they do so by studying the impact of organism level learning / behavior on evolution \parencite{Hinton1996, Belew1990, Gruau1993, Nolfi1999, Nolfi1994LearningAE, Whitley1994, Mayley1996, Turney2002, Bull1999, French1994, CarseO00, Parisi19911L, Ku2006}. \\
In real biology, problem-solving (intelligence) occurs not just in the 3D space of behavior, but in other subspaces as well (transcriptional, physiological, morphological etc..) \parencite{Fields2022}.\\

Here, we particularly focus on the intelligence of living systems in navigating anatomical morphospace. Developmental morphogenesis is a key stage in multi-cellular biology ensuring the self-organization of a cellular collective into a functional structure. Single cells, which participate in the collective traversal of morphospace during development, were once unicellular organisms themselves with many competencies \parencite{Lyon2015, Lyon2021}. It is unlikely that these competencies disappeared during the transition to multicellulalrity; evolution may have instead repurposed their competencies to serve anatomical goals. Clearly, this natural tendency for cells to communicate and rearrange themselves, must have implications to the evolutionary process.\\

Morphological development, and its subsequent impact on evolution is all but ignored in the field of evolutionary computation. Popular works consider the genotype to map directly to the phenotype (genotype $\rightarrow$ phenotype model) often ignoring the inverse relationship. We develop the more biological, genotype $\rightarrow$ mophogenesis $\rightarrow$ phenotype model, and study how the intelligence of the morphogenesis process impacts evolutionary dynamics in silico.

\section{Research Motivation}

Assessing the impact of morphogenesis on evolution can further our understanding in Evolutionary Computation and Developmental Biology: 

\begin{itemize}
    \item If morphogenesis proves beneficial to evolutionary search, it can be incorporated in standard genetic algorithms: genomes which act as subjects of evolution can be separated from the functional outcome by means of a morphogenetic process. Such a framework could help us grow functional morphologies with the distinctive robustness and plasticity seen in living systems.

    \item Understanding how evolution regulates morphogenesis could help shed light on confounding morphological phenomena in Biology. For instance, Planaria are a class of flatworms with a messy genome -- they have different chromosomes in each cell. Despite this, each fragment of Planaria regenerates to a complete worm with remarkable accuracy. How does an organism with the most chaotic genome have the best regenerative fidelity, immortality and cancer resistance ? 
\end{itemize}

\section{Research Question}

The main question we address here is:

\begin{itemize}
    \item[] How do diverse competency levels of the morphogenetic process impact the rate and course of evolution?
\end{itemize}

\section{Research Objectives}

\begin{itemize}
    \item Develop a modular software framework for the morphogenetic manipulation of artificial embryos.

    \item Introduce the developmental process as a novel step in the iterative sequence of a standard genetic algorithm.

    \item Develop a global fitness function capable of assessing the morphological state of an artificial embryo.

    \item Quantitatively assess the various ways in which degrees of competency of the developmental material (sitting between genotype and fitness) augments standard evolutionary search compared to passive (hardwired) architectures.

    \item Examine the role of hyper-parameters on the dynamics of evolution.

    \item Discuss limitations of the proposed framework.
\end{itemize}

\section{Main Contributions}

We study the impact of diverse degrees of morphogenesis on the evolutionary process by simulating the evolution of artificial embryos in silico. We introduce a new step -- morphogenesis, in the iterative sequence of a standard genetic algorithm. Morphogenesis allows cells of the artificial embryo to rearrange itself prior to selection. Morphogenesis can manifest in several degrees, with the least causing no change in embryonic structure, and the maximum causing the most change, resulting in perfect anatomical structure, as measured by the fitness function of our genetic algorithm. Embryos in our framework are cellular collectives, and we seek to understand  how the goal-directed, local, self-organization process carried out by individual cells during morphogenesis impacts evolution.\\

Results from our simulations reveal the extent to which morphogenetic-competency impacts evolution, and provides explanations for the confounding nature of Planaria. We discuss limitations, and possible extensions to our methodology, highlighting the applicability of developmental morphogenesis to the fields of Artificial Life and Evolutionary Robotics.

\section{Outline}

\begin{itemize}
    \item Chapter \ref{Chapter2}: \textbf{Background and Related work:} Provides our rationale to study the impact of morphogenesis on evolution. Discusses recent work.\\

    \item Chapter \ref{Chapter3}: \textbf{Methods:} Presents our framework to evolve artificial embryos using a genetic algorithm and defines terminology.\\

    \item Chapter \ref{Chapter4}: \textbf{Results:} Presents results from four experiments.\\

    \item Chapter \ref{Chapter5}: \textbf{Discussion:} Provides commentary on results obtained; discusses the limitations of our model and addresses future work.\\

    \item Chapter \ref{Chapter6}: \textbf{Conclusion.}
\end{itemize}

\chapter{Background and Related work} % Main chapter title

\label{Chapter2} % Change X to a consecutive number; for referencing this chapter elsewhere, use \ref{ChapterX}

\section{Background}

\subsection{Evolution and Intelligence, a Bidirectional Ballet}

\subsubsection{The Forward Relationship: Evolution Improves Intelligence}

The history of any living species is one of subsequent improvement. Looked at from a mathematical perspective, it is a process of optimization carried out to improve the reproductive fitness of living systems. Evolution by natural selection \parencite{darwin1859} is the algorithm which carries out such optimization in nature. \\

Inspired by the functional diversity of living creatures, the field of evolutionary computation seeks to replicate biological evolution to solve computational problems in the fields of economics, mathematics, engineering, and computer science. A major result of these efforts, and consequently a prime subject of discussion in this work, is the genetic algorithm (GA) by Holland \parencite{Holland1992}. Holland's main goal was to understand "adaptation" as it occurs in nature and find ways to bestow computers with the ability \parencite{Holland1975}.\\

A standard GA serves to improve populations of chromosomes -- points in solution space which map to outputs (the phenotype) -- from generation to generation. Chromosomes in a specific generation are assessed by a fitness function \parencite{Michalewicz1996} and only those chromosomes which perform well are selected forward to the next generation. \\

Algorithmically, given a population of chromosomes, the classic GA is an iterative sequence of the following steps: 

\begin{itemize}
    \item \textbf{Selection}: Each chromosome is assessed based on its task-solving capability by a fitness function. A fraction of the population which solve the task best are chosen forward to make a new population, while the rest are discarded.

    \item \textbf{Crossover}: The selected chromosomes, i.e., the parent chromosomes, repeatedly recombine with each other, to give rise to child chromosomes. The process terminates once the population regains its original strength. 

    \item \textbf{Mutation}: Chromosomes are stochastically modified to a random value so that a population does not stagnate in diversity. 
\end{itemize}

Genetic algorithms (GAs) are applied to a wide range of problems (for a detailed overview see \cite{Katoch2020}). An important consideration, well established in real biology, but often forgotten in works employing the GA, is that while evolutionary optimization can help improve problem-solving ability (i.e., intelligence), the degree of problem-solving ability itself can significantly influence evolutionary dynamics. The tendency to ignore the inverse relationship deprives employers of the GA from leveraging key ideas from Developmental Biology.

\subsubsection{The Inverse Relationship: Impact of Intelligence on Evolution}

In 1896, James Baldwin, an American psychologist, put forth a theory now known as the Baldwin effect, which hypothesized that adaptive behavioral traits acquired by a biological organism over the course of its lifetime can get genetically canalized into its genome over evolutionary timescales \parencite{Baldwin1896}. Independently, in 1953, Conrad Waddington, a British developmental biologist, produced in-vitro experimental evidence of a phenomenon similar to the Baldwin effect. Waddington showed that environmentally induced morphological changes can get genetically hardwired into a population, provided the factor inducing the change remains constant over evolutionary timescales in the environment \parencite{Waddington1953}.\\

In addition to being in accordance with real biology, the Baldwin and Waddington effects imply that a separate adaptive-process, acquired phenotypically by an individual over the course of its lifetime, can alter the genomic traits valued by evolution over evolutionary-timescales, potentially re-routing, and re-structuring the solution space towards optimal genomes.\\

In the late $20^{th}$ century, computer scientists, particularly those who sought artificial intelligence, took special interest in biological evolution. Inspiration from biology had then led to the artificial neural network model (NN) \parencite{McCulloch1943}, and enthusiasm to capture adaptive-biological-capabilities was at an all-time high. GAs had been successfully employed to optimize neural networks \parencite{montana1989training, whitley1990}, but were inferior to backpropagation (BP) \parencite{Rumelhart1986}. The issue was that GAs failed to navigate evolutionary search spaces as well as BP did, sinking into premature local optima \parencite{Yao1997}.\\

Geoffrey Hinton and Steven Nowlan, then computer scientists at Carnegie-Mellon, sought to understand why biological evolution didn't suffer from this problem. Their key insight came from the Baldwin effect, namely, lifetime-behaviors of an individual could act as local search-subroutines, guiding global evolutionary search towards optimal solutions. In their experiments, they evolved populations of neural networks (NNs) with a GA, but each NN executed a learning-subroutine before selection. Results from these simulations revealed that NNs with learning could evolve much faster than those without \parencite{Hinton1996}. The impact of learning on evolution was the first in-silico evidence of the Baldwin effect, and consequently that of the inverse relationship as well. 

\subsection{A Re-framed Perspective of Intelligence}

The inverse relationship has since received considerable attention from the computer science community. Unsurprisingly so, because gains in the evolution of intelligence (forward relationship) can be compounded by leveraging the effect of intelligence on evolution (inverse relationship)\parencite{Belew1990}. By "intelligence", here, we mean the ability of an entity to solve pre-defined tasks. \\

The traditional way of studying the impact of intelligence on evolution has been to define it in terms of a learning process. Learning has been treated as a substitute for behavior, adaptation, or any other phenomenon observed during the lifetime of a biological organism \parencite{Hinton1996}. This overview treats intelligence as a set of observable abilities in the three dimensional (3D) physical world. However, this is but one domain in which intelligence manifests.\\

\begin{figure}[htbp]
\includegraphics[width=14cm]{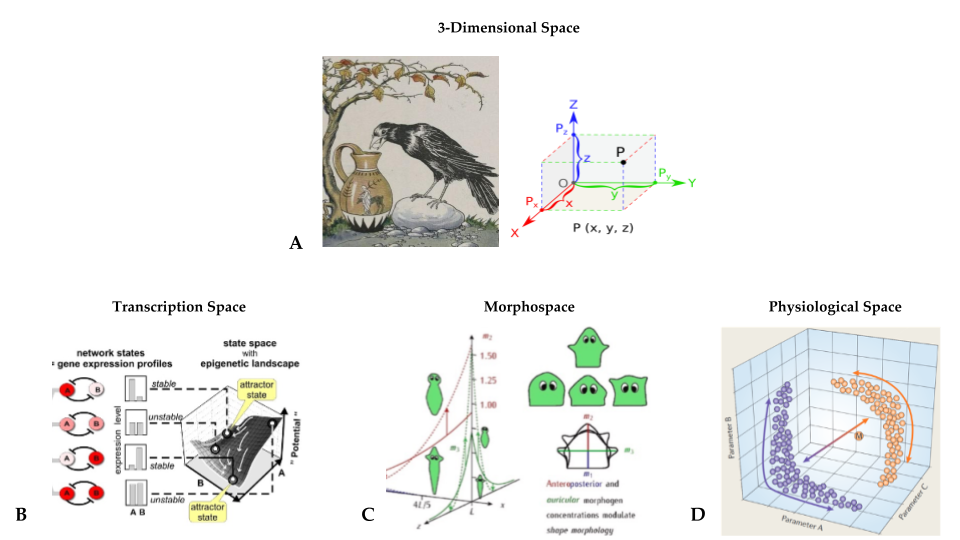} 
\centering
\caption[Living Systems exhibit Intelligent Problem-solving in Multiple Spaces]{Living systems exhibit intelligent problem-solving in multiple spaces, taken with permission from \parencite{Fields2022}. \textbf{(A)}: Task solving ability of an agent in the physical 3D world is readily seen as intelligence: observing a bird drop rocks into a partially filled water-jug hints at its ability to solve the problem of thirst by employing intuitive physics. The bird navigates a space of possible solutions in three-dimensional world (right panel) by moving its body and the stone to an optimal solution. \textbf{(B)}: Problem solving in transcription space: the space of all possible gene expression combinations. An example of a gene-regulatory network between two genes A,B and its corresponding epigenetic landscape is shown. The state of gene A influences B and vice versa. Each combination of expression maps to a point in 2D epigenetic space. Navigation through this space corresponds to sampling gene-combinations. \textbf{(C)}: Problem solving in morphospace: the space of all possible structural anatomies. Planarian head shapes are determined by expressing morphogens in different combinations (three shown here based on results from a computational model). \textbf{(D)}: Problem solving in physiological space: the space of all possible physiologies. An example with three variables (parameters A, B and C) is shown, these parameters can be, for instance, ion concentrations. An organism has to pick the correct physiology based on its environment.} \label{figure-multidomainIntelligence}
\centering
\end{figure}

Living systems have the capability to solve problems not just in the conventional 3D space around us but in other arbitrary spaces as well. This includes transcription, physiological, and anatomical sub-spaces \parencite{Fields2022}. As observers, us humans (and indeed, other animals), are by default adept at recognizing intelligence in the physical 3D world: observing a bird drop stones into a water-filled jug, hints at its ability to solve the problem of thirst by using intuitive-physics -- it is capable of navigating the space of solutions in the 3D world (Figure \ref{figure-multidomainIntelligence}A).  Similarly, we are also capable of recognizing agency at the higher dimensional level of social interactions: a group cooperating to achieve a goal can be said to navigate solutions in a space of "who-does-what-now". However, we are poor when it comes to recognizing agency in unconventional spatio-temporal scales and embodiments.\\

For instance, gene-networks constantly process and manipulate information from multiple genes to solve the task of gene expression. These dynamic systems navigate the space of all possible gene-combinations (transcription space) to isolate a single relevant solution (Figure \ref{figure-multidomainIntelligence}B). Clearly, this non-random search process is a form of intelligence, but we wouldn't readily recognize it so because it lies in a different domain (the transcription space) from the one we are familiar with. Similarly, living systems can navigate the space of all possible physiologies towards the one correct for its environment (Figure \ref{figure-multidomainIntelligence}D).\\

\begin{figure}[htbp]
\includegraphics[width=12cm]{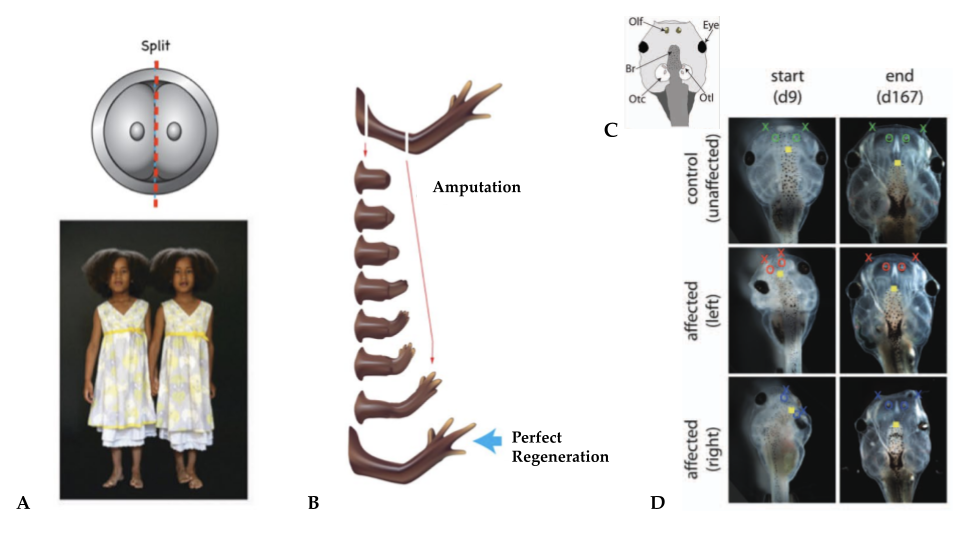} 
\centering
\caption[Intelligence of Developmental Morphogenesis]{Competency of developmental morphogenesis to reliably navigate through anatomical morphospace despite a high degree of perturbation \textbf{(A)}: Most mammalian embryos when cut into two jagged parts develop into normal sized monozygotic twins despite drastically different cell counts in the split embryos. \textbf{(B)}: Regenerative animals such as axolotl’s are capable of growing an amputated limb reliably towards a target morphology, despite different starting states. Cellular-collectives might be executing a coordinated process of re-growth to ensure such reliable anatomy; taken with permission from \parencite{Fields2022} \textbf{(C)}: Facial features of a normal tadpole (dorsal view) (br: brain; olf: olfactory bulbs (nostrils); Otc: otocyst, otl: Otolith). \textbf{(D)}: Scrambled craniofacial structures of a tadpole reliably develops into a largely normal frog face \parencite{vandenberg2012normalized}. First Row: craniofacial features of a normal tadpole on the ninth day (d9) post fertilization (left panel) and the completely developed face on the 167th day (d167) post fertilization (right panel). Second Row: craniofacial organs perturbed to abnormal positions on the left embryonic hemishphere (left panel), and its subsequent development on d167 (right panel). Third row: craniofacial perturbation on the right embryonic hemisphere (left panel) and subsequent development (right panel). Facial organs move through novel pathways during the course of development to from largely normal frog faces at the end of development. Given that cells of such a multi-cellular collective were once uni-cellular organisms with complex context sensitive capacities \parencite{Lyon2015}, their communication with one another may have led to their coordination towards the correct morphological outcome despite extreme perturbation; taken with permission from \parencite{vandenberg2012normalized}} \label{figure-morphospace}
\centering
\end{figure}

This pattern of problem solving extends to the domain of anatomical morphospace as well, and is of particular importance to our discussion (Figure \ref{figure-morphospace}). A rich set of examples support morphogenesis as an intelligent, goal-directed homeodynamic process. The most striking examples, to this end, are seen in regulative development and regeneration \parencite{pezzulo2016, Harris2018, goss2013principles} where cells work to implement and maintain a large-scale form (target morphology) despite surgical, genetic, and physiological sources of defects: many mammalian embryos when split into two unequal parts proceed to develop into normal whole bodies (monozygotic twins) making up for very drastic damage to nevertheless achieve correct morphology \parencite{Beoluk2020} (Figure \ref{figure-morphospace}A). Regenerative organisms, such as Planaria and Axolotl’s can regrow parts of their body exactly to a target morphology, with re-growth neither terminating prematurely nor proceeding perpetually, despite multiple starting positions of the handicapped part \parencite{goss2013principles} (Figure \ref{figure-morphospace}B).\\ 

An extreme example of anatomical homeostasis, and one inspiring our framework, concerns the Picasso like scrambling of the facial organs of a tadpole during embryonic development, which, despite highly dislocated positions, proceed to move through novel paths during development to form normal frog faces \parencite{vandenberg2012normalized} (Figure \ref{figure-morphospace}C,D). The ability of craniofacial organs to move in such a manner requires careful coordination between not just its own cells but that of other somatic cells as well. This would have been unlikely if cells were passive materials executing tasks as directed by a pre-defined rule set (such as the genome). Indeed, individual cells within such a multi-cellular collective having once been uni-cellular organisms themselves with complex capabilities for sensing and navigation, are not passive, but agential materials \parencite{Lyon2015, Lyon2021}. Evolution could have repurposed their agential capacities to ensure self-organized behaviour towards a common goal \parencite{fields2020morphological}.\\

Given the role morphogenetic competency plays in reorganizing cells towards an anatomical outcome, it must have an important role in influencing evolution. Here, we study if, and to what extent, such reorganization capacity grants evolving populations an advantage over those who cannot.

\section{Related Work}

The role of morphogenesis on evolution has so far not been addressed in the field of evolutionary computation. The latest explored, has been the impact of 3D behavior, specifically learning, on evolution, spearheaded by Hinton and Nowlan \parencite{Hinton1996}. Here we review works assessing the inverse relationship post their seminal work.\\

Several works provided confirmatory analyses of the effect of learning on GA evolution \parencite{Belew1990, Gruau1993, Nolfi1994LearningAE, Nolfi1999}. Darrell Whitley compared function optimization by classic GAs vs. GAs with learning, discovering that learning significantly improved GA optimization \parencite{Whitley1994}. The indication that learning could improve evolution by GAs, pushed the community towards better understanding the conditions under which learning impacted evolution. A positive correlation between the phenotypic and genotypic spaces was found to be necessary for learning to have a beneficial effect \parencite{Ku2006}. Irrespective of the correlation between the local-search process of learning and the global search of the fitness function, learning seemed to improved evolutionary search \parencite{CarseO00, Parisi19911L}. Inclusion of costs to the learning process encouraged genetic assimilation of beneficial traits, leading to the conclusion that a cost-benefit trade-off was responsible for the Baldwin Effect \parencite{Mayley1996, Turney2002, Mayley1996Conditions}. The rate, amount of learning \parencite{Bull1999}, and manner of crossover \parencite{French1994, fontanari1990effect} employed in the GA were all found to have an impact on evolution. \\

Knowledge from these studies led to their adoption in evolutionary algorithms, bringing about a drastic increase in GA performance \parencite{Whitley1994}. The reason for this improvement circles back to Hinton and Nowlan’s work. By simulating the Baldwin effect, the authors inadvertently introduced a non-linear component between the genotype and phenotype. Most of the field until that point applied GAs assuming a direct map between the genotype and the problem-solving-capability (phenotype) -- the genotype $\rightarrow$ phenotype model. Introduction of the local search process, in the form of learning, disrupted this direct map, essentially bringing in a non-linearity. The non-linear relationship between the genotype and the phenotype not only made GA evolution more biological but made it possible to study the inverse relationship, as we explain below.

\subsection{Learning Restructures the Evolutionary Landscape}

The traditional evolutionary landscape is a space of genotypes, determined by the genomes of a population in any generation. Such a landscape is riddled with peaks / valleys owing to the diversity of the population. In the absence of a nonlinearity between the genotype and the phenotype, the classic GA finds it notoriously difficult to navigate such a complex space, often prematurely settling in local optima \parencite{mitchell1995genetic, Whitley1994}. A complementary local search process, such as learning, can guide evolution through this rugged landscape \parencite{Hinton1996}. Selection can focus on getting to the next major peak / valley and learning can navigate around this region towards a better optima), essentially restructuring the fitness landscape for evolution.\\

Consequently, the Genotype $\rightarrow$ Learning $\rightarrow$ Phenotype model, encompasses two fitness landscapes: 1. the original, genotype determined landscape, and 2. the learning altered, phenotype-based landscape. By altering the type/amount of learning, one could visualize the amount by which the phenotypic landscape diverged from the genotypic landscape, helping quantize the effect learning had on evolution (nothing but the inverse relationship).

\subsection{The Genotype - Morphogenesis - Phenotype Model}

Here, we step aside from learning, and assess how different degrees of problem-solving competencies of the morphogenetic process diverts the phenotypic landscape away from the genotypic landscape, and how the divergence effects the genotypic landscape itself. To this end, we carry out the evolution of artificial embryos in-silico using a classic GA \parencite{Holland1992}. \\

In our model, we add morphogenesis as the only extra step to a classic GA. Inspired by the picasso like scrambling of tadpole facial organs, and its subsequent reorganization towards a normal frog face over the course of development \parencite{vandenberg2012normalized}, we design our morphogenetic process to reorganize an initially disorganized jumbled mess of cells into a target morphological order. Consequently, each “embryo” in our framework is a multi-cellular collective. We provide embryos with different degrees of reorganization capacity and observe how each level of competency affects selection by a GA tasked with picking embryos with the most monotonic order. No external environment is considered. \\

We provide a detailed description of our framework in the next chapter.
 
% Chapter Template

\chapter{Methods} % Main chapter title

\label{Chapter3} % Change X to a consecutive number; for referencing this chapter elsewhere, use \ref{ChapterX}
The following chapter borrows significantly from previously published work \parencite{Shreesha2023}. Respective contributions are listed in the contributions section.\\

We simulate the evolution of artificial 1-dimensional embryos in silico. The following sections describe the structure of each embryo, our paradigm for modeling developmental morphogenesis towards a target adult anatomy, and the process of selection employed to study their dynamics over time.

\subsection{Creating Populations for Evolution}

A population consists of a number of embryos. Each embryo is represented as a one-dimensional array of fixed size (matching the cell count in the 1-dimensional embryo). Each cell of this array is initialized with a different integer value representing the positional value "gene" for the corresponding cell of the embryonic axis. In this minimal model, there is no further chromosomal structure or transcriptional change, and we simply refer to the structural genes as directly specifying the positional preference of a given cell. Each embryo undergoes a developmental cycle (described below) to become a mature “individual”. We model evolution in three kinds of populations: a “hardwired” population consisting of only hardwired embryos, a “competent” population of only competent embryos, and a “mixed” population which contains both kinds of embryos, in varying proportions. For most simulations reported here we chose a population size of $n = 100$ embryos. The one exception is in the simulations of hybrid populations, for which $n = 200$.

\subsection{Hardwired and Competent Embryos}

\begin{figure}[htbp]
\includegraphics[width=12cm]{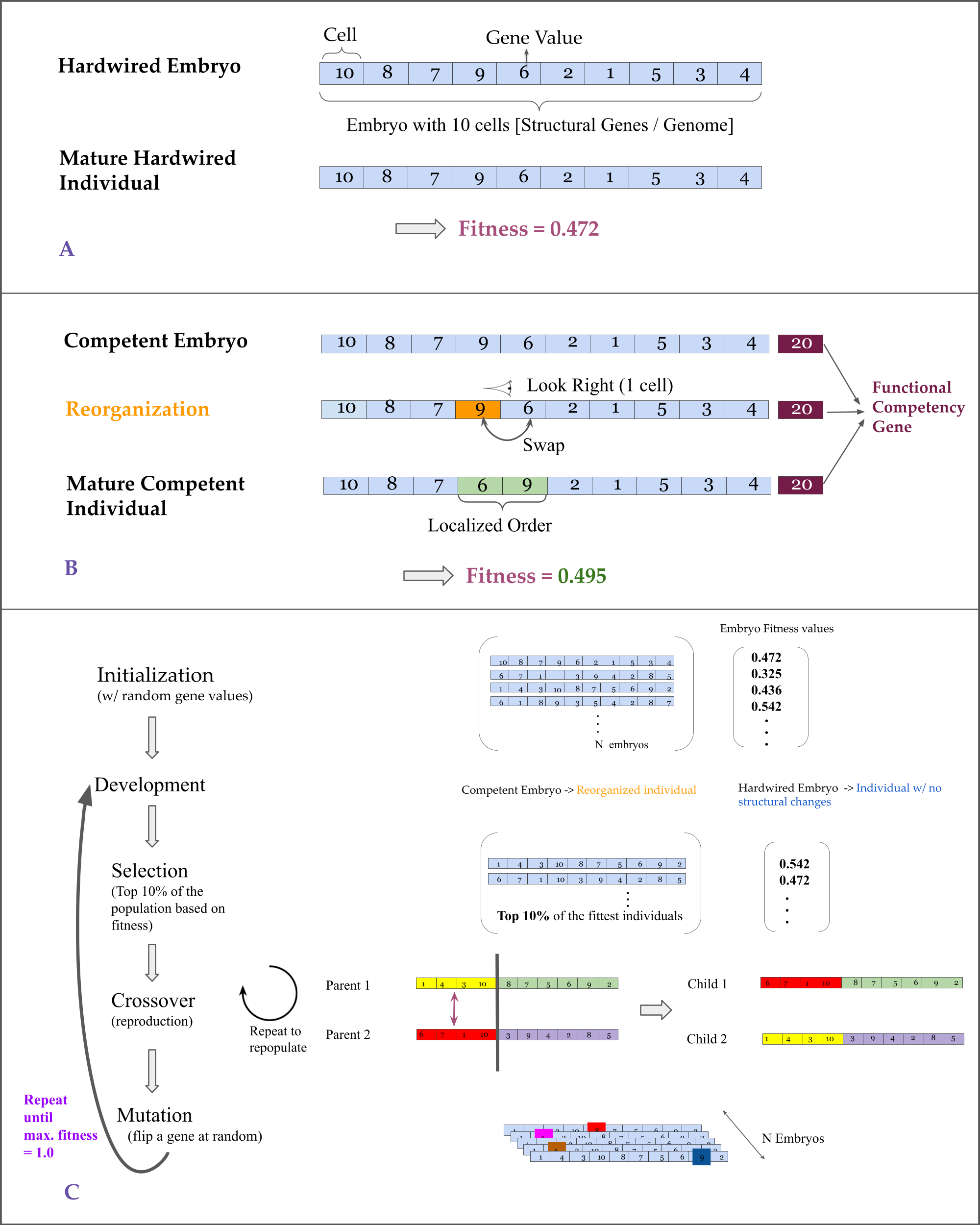} 
\centering
\caption[Schematic of Experimental Setup]{Schematic of experimental setup. Figure redesigned from \parencite{Shreesha2023} \textbf{(A)}: Definition of a hardwired embryo: each hardwired embryo is a 1-D array consisting of 50 cells (10 shown here as example). Each cell takes an integer value between $[1, 50]$, and is considered to be its Structural Gene. The fitness of an individual is defined as the degree of order within its genes (0 implying descending order, 0.5 implying random order and 1.0 implying ascending order). In the example shown here, the embryo is randomly initialized and hence has a fitness close to 0.5. \textbf{(B)}: Definition of a competent embryo: Each competent embryo is identical to a hardwired embryo except that it carries an additional functional “competency gene” indicating how many cell movements it can carry out during a developmental cycle to achieve ordered ascending arrangement before phenotypic assessment. The functional gene can be locked down to a pre-specified value for an entire population or can be made evolvable. \textbf{(C)}: Description of the genetic algorithm used to evolve hardwired and competent embryos. See methods (chapter \ref{Chapter3}) for details }\label{figure-mainMethods}
\centering
\end{figure}

We define two types of embryos, a “hardwired embryo” and a “competent embryo” (Figure \ref{figure-mainMethods} A,B). The difference between them lies in the way they develop during the evolutionary cycle. A competent embryo consists of cells capable of sensing neighboring cells and adapting morphology by moving around locally prior to the adult stage in which fitness is evaluated. “Competency” is the capability of these embryos to carry out such reorganization, and they carry a gene (competency gene) that dictates the frequency at which they can do so (fixed, in some experiments, but free to evolve in others). Our competent embryos leverage sensing and motility to reorganize their cells during ‘development’ in a way that boosts fitness (see below). We vary the degree to which they can reorganize themselves (competency level). A hardwired embryo lacks this capability; its structure from birth to maturity is constant.

\subsection{Development: Restricted Bubble Sort}

At the beginning of each evolutionary cycle, competent embryos are considered “just born”, their morphological structure having been decided by their parents from the previous generation. Therefore, their fitness at this point of time is called the “genotypic fitness”. Hardwired embryos and competent embryos both carry a genotypic fitness at the start of every evolutionary cycle. Soon after, embryos undergo a process of development. Competent embryos carry out restricted-bubble sort to rearrange their cells in a way to boost fitness (i.e in a way to increase ascending order of its elements). To do so, each competent array carries an extra value known as the competency value. This integer determines how many successive bubble-sort swaps will take place during its developmental cycle. A higher competency value implies the ability of a competent embryo to reorganize its structure to a greater degree, and vice-versa. Usually, competency levels are much lower than the total number of bubble-sort swaps required by an embryo to attain maximum fitness, for this reason it is called “restricted” bubble-sort. Hardwired embryos have no such reorganization capability. They end their life cycle with the same structure as that of at birth. \\

At the end of their respective developmental cycles, embryos become “individuals”: competent embryos become competent individuals, and hardwired embryos become hardwired individuals (even if nothing changes structurally in them). At this point, the monotonicity of each embryo’s array is calculated again to determine the “phenotypic fitness” of the individual. Since competent “individuals” have rearranged cells by bubble sort during development cycle, their phenotypic and genotypic fitnesses diverge. In contrast, hardwired individuals do not rearrange, therefore their genotypic and phenotypic fitnesses are identical. We assess the impact of the morphogenetic process on evolution by observing the divergence between these two fitness curves.\\

Algorithm \ref{Alg2: development} provides an overview of our developmental process.

\begin{algorithm}

 \caption{Development Function} \label{Alg2: development}
 \SetKwFunction{FMain}{development} 
  \SetKwProg{Fn}{Function}{:}{}
  \Fn{\FMain{competentPopulation, competencyValue}} {
  
        \For {embryo $\in$ competentPopulation} {
        
        individual $\leftarrow$ BubbleSort(embryo, competencyValue)\;
        comepentPopulation.replace(embryo, individual)\;     
        } 
    }      
\KwRet competentPopulation\;
\end{algorithm}

\subsection{Fitness of Embryos and Individuals}
We define fitness as the degree to which an \textit{individual’s} array of integers is in ascending order. Individuals with cells arranged in ascending order by value are attributed a fitness of 1.0 (maximum), those whose cells are randomly ordered are attributed a fitness of 0.5. We calculate the fitness (the degree of order) of an array by counting the number of non-inversions present in it:\\

Consider an embryo (a one-dimensional array) $\textbf{A}$, of size $\textbf{n}$ initialized with random integer values in the range of $[1, n]$. Let $A(0), A(1), \dots , A(n)$ be its elements.\\ 

The “non-inversion” count of this array ($nIC$) is a count of the number of elements which \textit{do not} require to be swapped for the array to have ascending order. Specifically,

\begin{equation} 
    nIC(A) = \# \hspace{0.1cm}\{ (A[i], A[j]) \hspace{0.2cm} \mid \hspace{0.1cm} i<j \hspace{0.2cm} \& \hspace{0.2cm}A[i]<A[j] \} 
\end{equation}

where, 
$i \neq j$, \\
$i = 0, 1, \dots, n-1$ \\
$j = 1, 2, \dots, n-1$\\

we normalize the non-inversion count ($nIC$) of array $A$ as follows:

\begin{equation}
    nIC^{'} = \frac{nIC(A)}{\Comb{n}{2}}
\end{equation}

and report fitness ($f$) on an exponential scale so as to "zoom into" higher fitness values:

\begin{equation}
    f = \frac{9^{nIC^{'}}}{9}
\end{equation}

\subsection{Genetic Algorithm}
To evolve populations (hardwired or competent), we employ a genetic algorithm (GA) with the following steps (Figure \ref{figure-mainMethods}C and Algorithm \ref{Alg1: GeneticAlg}):
\begin{itemize}
    \item \textbf{Development}: Post initialization, embryos undergo development. A competent embryo reorganises its cells based on its competency value, whereas a hardwired embryo does not. 

    \item \textbf{Selection}: The fittest 10\% of individuals in a population are selected to move on to the next generation. Selection in a population is based on its individuals’ phenotypic fitness (Darwinian selection). 

    \item \textbf{Cross-Over}: In order to repopulate a population back to its original strength, we carry out a process of reproduction called cross-over. It occurs as follows: Two individuals are involved, each of these are split at a random location along their length. One half of Individual 1 is swapped with the same half of Individual 2 to give rise to two children. Figure \ref{figure-mainMethods}C contains an illustration of this process.

    \item \textbf{Mutation}: The repopulated population is subjected to random point mutations. We set the probability of an individual receiving a point mutation to be 0.6.
\end{itemize}

\begin{algorithm}

 \caption{Genetic algorithm} \label{Alg1: GeneticAlg}
 \SetKwFunction{FMain}{Main} 
  \SetKwProg{Fn}{Function}{:}{}
  \Fn{\FMain{population, competencyValue}}{
    \While {bestFitness < 1.0} {
        ReorgPopulation $\leftarrow$ development (population, comptencyValue)\;
 
        selectedPopulation $\leftarrow$ selection(ReorgPopulation)\;
        
        RepopulatedPopulation $\leftarrow$ crossover(SelectedPopulation)\;
        
        population $\leftarrow$ mutation(RepopulatedPopulation, probability = 0.6) \;

        bestFitness $\leftarrow$ maxFitness(population) \;
 
        } 
  }
  \KwRet\;
\end{algorithm}

% Chapter Template

\chapter{Results} % Main chapter title

\label{Chapter4} % Change X to a consecutive number; for referencing this chapter elsewhere, use \ref{ChapterX}

The following chapter borrows significantly from previously published work \parencite{Shreesha2023}. Respective contributions are listed in the contributions section.\\

We built a virtual embryogeny model in which fitness was defined by the degree of monotonicity of a 1D array of numbers, simulating a minimal organism with a single axis of positional information (Figure \ref{figure-mainMethods}). The initial sequence of numbers for each embryo was assigned randomly. Since these sequences decided the embryo’s structure (cell order), they are referred to as its "structural genes". As described in Chapter \ref{Chapter3}, the structural gene sequence of hardwired embryos is fixed: their genome directly encodes for their phenotype. For competent embryos, we implemented different degrees of competency during a period of `development' during which cells were allowed some degree of movement relative to their neighbors, allowing them to reorganize to improve monotonicity prior to evaluation of phenotypic fitness. This enabled phenotypic fitness for competent individuals to diverge from raw genotypic fitness, with the extent of divergence depending on how much cell movement was permitted. This corresponds to different degrees of capacity for cells in-vivo to optimize homeostatically preferred local conditions with respect to informational signals such as positional cues and polling of neighboring cell states. An evolutionary cycle was implemented around these developmental events using a GA. \parencite{Holland1992}.\\

In initial experiments, the competency gene was fixed to a pre-determined value across the evolutionary run, enabling study of evolutionary dynamics over time as a function of different degrees of morphogenetic competency.

\subsection{Morphogenetic Competency Accelerates Evolutionary Search}

\begin{figure}[htbp]
\includegraphics[width=10.9cm]{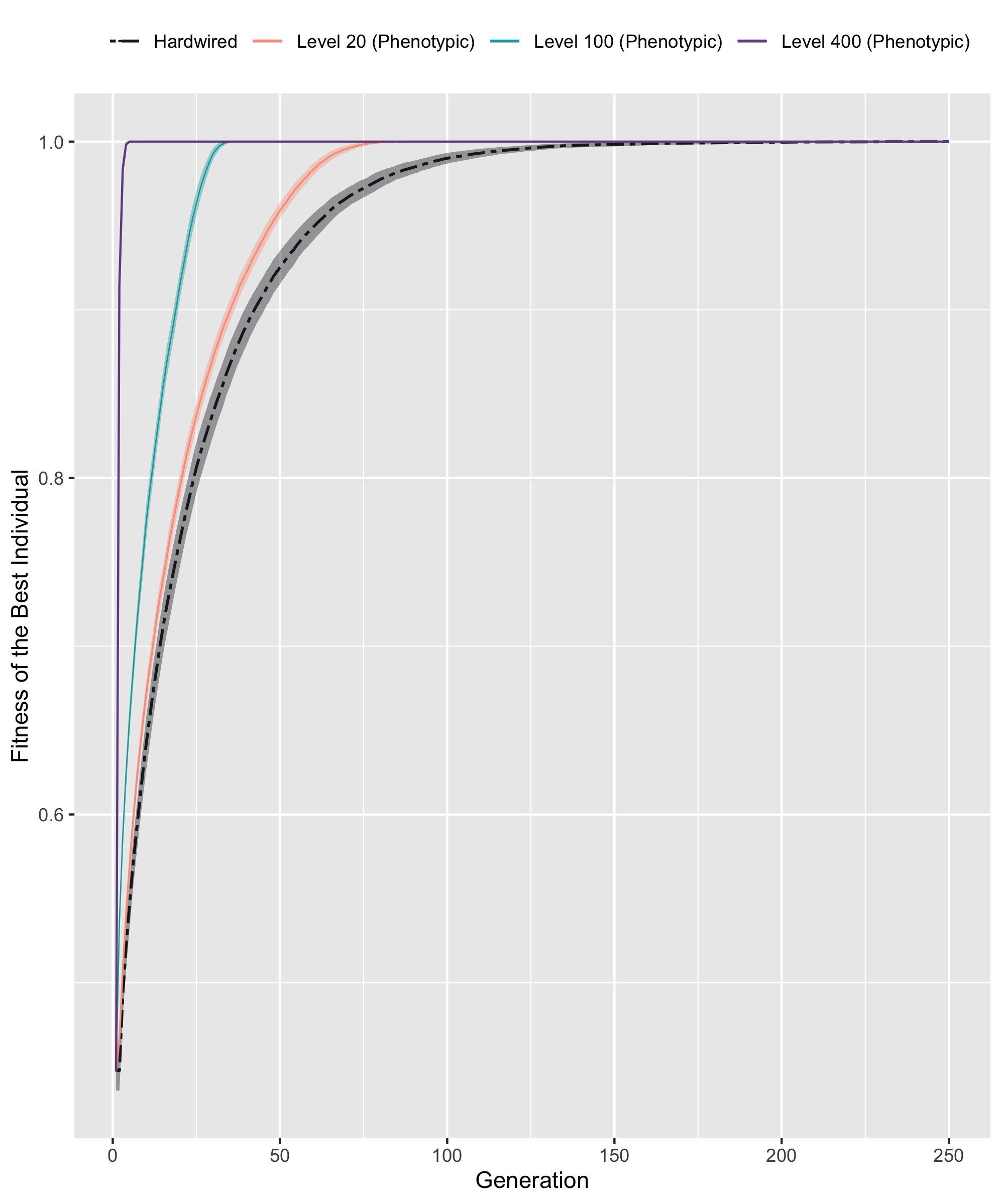} 
\centering
\caption[Experiment 1: Hardwired vs. Competent (Phenotypic Fitness Curves)] {Competent individuals have a higher rate of fitness than their Hardwired counterparts. Figure recreated from \parencite{Shreesha2023}. Three populations with different competency levels [Levels 20, 100, and 400] and a single hardwired population were initialized. Competency level refers to the maximum number of cell-swaps a competent embryo can execute during its developmental cycle. The individual with the maximum fitness in each population was plot over 250 generations. Shaded areas represent 95\% confidence interval bands over $n = 100$ repeats of each experimental condition.} \label{exp1:pheno}
\centering
\end{figure}

We first compared, over 250 generations, the time-course of evolutionary search towards a fully ordered axis in hardwired vs. competent individuals. After 100 generations, the hardwired population had the least fitness compared to populations with varying degrees of competency (Figure \ref{exp1:pheno} and Table \ref{Table1:exp1Times}) Table \ref{Table1:exp1Times}, provides a summary of the generation number at which each population crossed different fitness thresholds. We compared fitness of the best individual in competent and hardwired populations at generations 2, 10, and 20 (because these points exhibited the greatest sample variances.) At each of these, the difference in fitness between hardwired and competent populations was significant (p-values << $1 \times 10^{-3}$ for all points, Student’s t-test; for details see Appendix \ref{AppendixExpDetails}).\\

\begin{longtable}{l|rrrrrr}
\caption[Population Break-Through Frequencies for Experiment 1]{The number of generations different populations take to break through a particular fitness threshold. Table recreated from \parencite{Shreesha2023}. The break-through times reported are for the best individual in the population. Competency level indicates the number of swaps available to each embryo when initialized.} \label{Table1:exp1Times}\\ 
\toprule
\multicolumn{1}{l}{} & \multicolumn{6}{c}{\textbf{Fitness Threshold}} \\ 
\cmidrule(lr){2-7}
\multicolumn{1}{l}{\textbf{Competency Level}} & 0.65 & 0.75 & 0.8 & 0.9 & 0.97 & 1.0 \\ 
\midrule
No competency (Hardwired) & 10 & 18 & 24 & 42 & 72 & 250 \\ 
Level 20 & 9 & 16 & 21 & 36 & 55 & 93 \\ 
Level 100 & 5 & 9 & 12 & 19 & 26 & 37 \\ 
Level 400 & 2 & 2 & 2 & 2 & 3 & 5 \\ 
\bottomrule
\end{longtable}

Figure \ref{exp1:pheno} also shows that the 95\% confidence interval bands over 100 repeat runs decreased with increasing competency level, suggesting that more competent architectures are also more consistent in performance over time. Note that hardwired individuals gradually improved to reach peak fitness, taking well over 200 generations to do so, whereas the most competent individuals (with a competency level of 400) did so in under 6 generations. These data demonstrate the role competency plays in non-linearly improving the rate of fitness of a population and supports a clear conclusion: the higher the competency, the better the performance.\\

\begin{figure}[htbp]
\includegraphics[width=10.9cm]{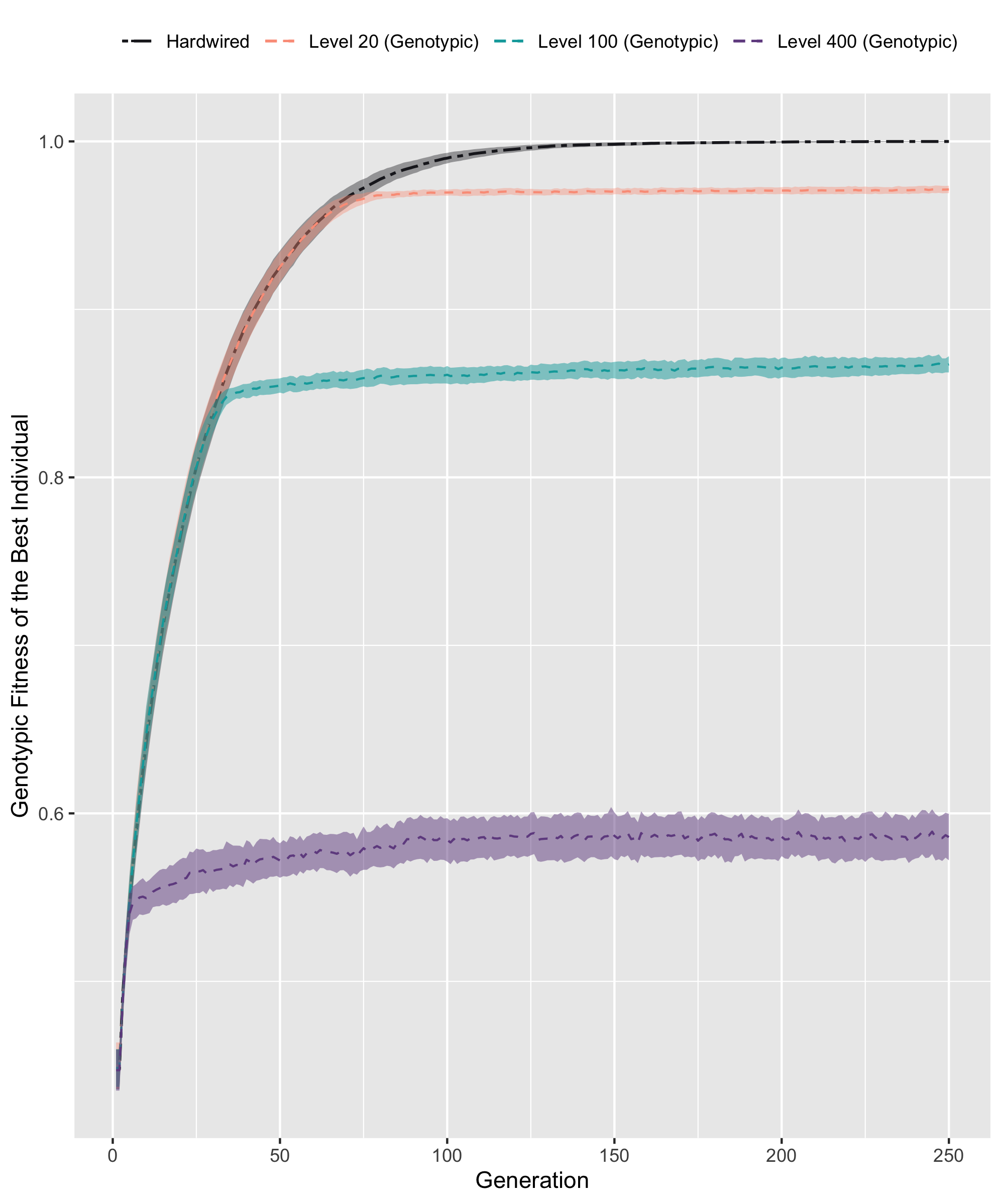} 
\centering
\caption[Experiment 1: Hardwired vs. Competent (Genotypic Fitness Curves)]{Competency comes at the expense of reduced genotypic fitness. Figure recreated from \parencite{Shreesha2023}. Genotypic fitnesses of the best individual in three different competent populations (competency levels 20, 100, and 400) were compared with that of a hardwired population over 250 generations. Genotypic fitness of a competent embryo was calculated prior to its development. Since hardwired embryos do not develop, they carry a genotypic fitness by default. Shaded areas in the figure represent 95\% confidence interval bands over $n = 100$ repeats.} \label{exp1:geno}
\centering
\end{figure}

Based on the impact of competency, one could hypothesize that progressively increasing competency would lead to a progressive decrease in selective pressure for good structural genes to appear. An embryo with high competency would have no selective pressure to improve its structural genes beyond a certain level because it can rely on its competency to re-order its cells to reach peak fitness. This is in fact what we observed (Figure \ref{exp1:geno}). We compared the genotypic fitnesses of the best individual in three populations with different levels of competency (20, 100, and 400) to that of a hardwired population. In all three competent populations, genotypic fitness rose with that of the hardwired population for a few generations, after which it plateaued, indicating that at this point, the structural genes were good enough for competency to achieve a phenotypic fitness that insured selection. Further, with increased competency, the 95\% confidence interval bands for genotypic fitness grew wider. Thus, as hypothesized, increasing competency in our simulation enabled excellent performance but reduced selective pressure on the embryo’s structural genes.\\

\subsection{Competent Individuals take over Mixed Populations}

\begin{figure}[htbp]
\includegraphics[width=12cm]{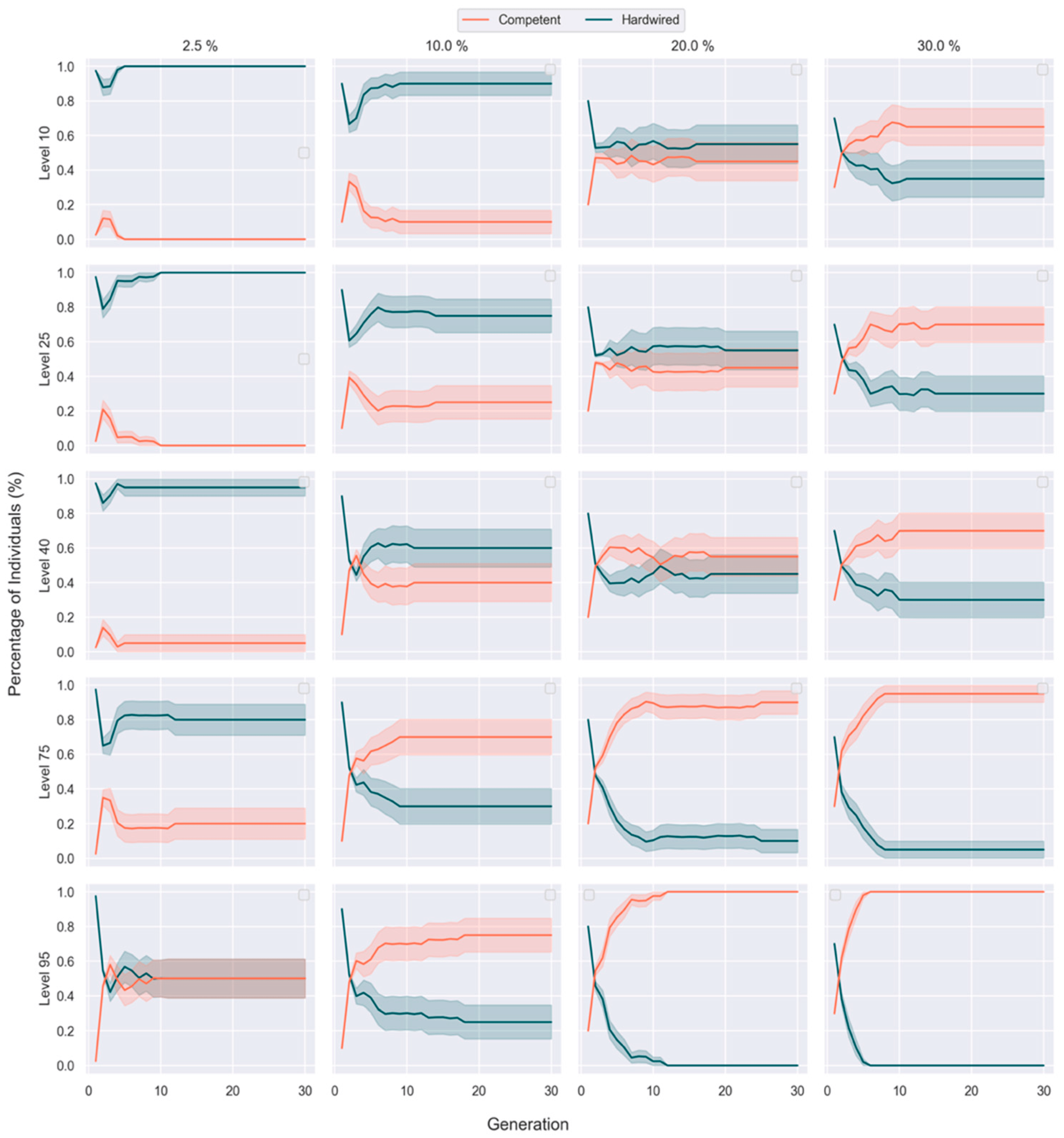} 
\centering
\caption[Experiment 2: Mixed Population Evolution]{Competent individuals dominate over hardwired individuals in a mixed setting when given adequate competency. Figure recreated from \parencite{Shreesha2023}. Each column represents the percentage of competent embryos in a hybrid population ($n = 200$ total) at initialization, increasing from left to right. Each row shows data from experiments at different competency levels, which increase from the top to bottom. Simulations were run for 30 generations. Shaded area represents variance over 20 repeat runs of each experiment} \label{exp2:mixed}
\centering
\end{figure}

Given these tradeoffs, we next asked how mixed populations (200 embryos per population) of competent and hardwired embryos would evolve (Figure \ref{exp2:mixed}). We varied both the level of competency and the percentage of competent embryos in the hybrid population at the start of the simulation. To probe the levels of competency required for embryos to dominate the population over the evolutionary simulation, competent embryos were always initialized as a minority of the starting population. Relationships between competency, initial population proportion, and dominance were observed over several runs.\\

When competent embryos constituted just 2.5\% of the initial population, they failed to dominate even at the highest level of competency tested: embryos with a competency level of 95 merely reached equal percentages with hardwired embryos. As their initial proportion in the population increased, competent embryos required progressively less competency to dominate over their hardwired competitors. At 10\%, embryos with a competency level of 75 could dominate; at 20\%, the competency level required for domination decreased to 40; and at 30\%, competent embryos dominated with a competency as low as 10 (Figure \ref{exp2:mixed}). In all starting conditions that resulted in dominance of competent embryos, it occurred rapidly, in just two or at most three generations (Table \ref{Table2: mixed domination}).

\begin{longtable}{l|lllr}
\caption[Domination Times for Mixed Population Evolution (Experiment 2)]{Time taken by competent embryos to dominate over hardwired embryos when mixed together in different ratios. Table recreated from \parencite{Shreesha2023}. Each column indicates the proportion of competent embryos in a hybrid population of size 200. The remaining embryos of the population are hardwired. Each hybrid population was evolved over 30 generations with a fixed level of competency (rows). Competent embryos are said to dominate when their prevalence rises over that of hardwired embryos and continues to rise or remains stable without dropping. Values indicate the number of generations required for competent individuals to dominate over hardwired individuals. “$\times$” indicates no dominance.}\label{Table2: mixed domination} \\ 
\toprule
\multicolumn{1}{l}{} & \multicolumn{4}{c}{\textbf{Percentage of Competent Embryos}} \\ 
\cmidrule(lr){2-5}
\multicolumn{1}{l}{\textbf{Competency Level}} & 2.5\% & 10\% & 20\% & 30\% \\ 
\midrule
Level 10 & x & x & x & 3 \\ 
Level 25 & x & x & x & 3 \\ 
Level 40 & x & x & 3 & 3 \\ 
Level 75 & x & 3 & 3 & 2 \\ 
Level 95 & x & 3 & 2 & 2 \\ 
\bottomrule
\end{longtable}

\subsection{Evolution Prefers a High, Constant Level of Competency}

\begin{figure}[htbp]
\includegraphics[width=12cm]{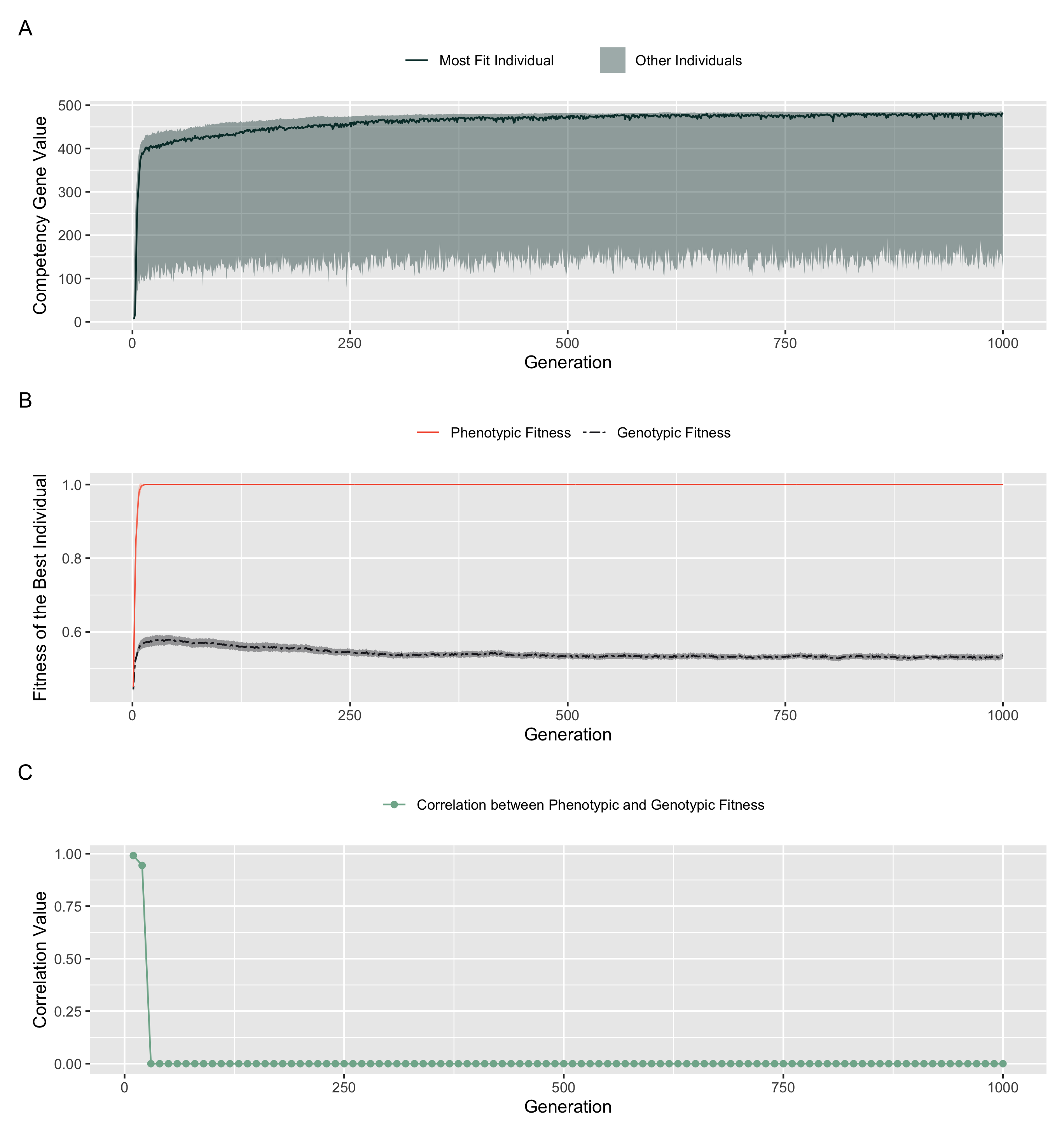} 
\centering
\caption[Experiment 3: Evolvable Competency]{Allowing evolution to set competency level: a perfect genome is not required to boost fitness. Figure recreated from \parencite{Shreesha2023}. Competency gene values for embryos ($n = 100$) were randomly initialized in the range of $[1, 15]$. Over the course of evolution each competency gene was allowed to mutate to a value in the range of $[1, 500]$. \textbf{(A)}: Competency gene value of the most fit embryo over the course of evolution. Shaded area represents the range of competency gene values in the population. \textbf{(B)}: Fitnesses of the best individual in a population of competent embryos with evolvable competency. Shaded area represents variance over 100 runs. \textbf{(C)}: Correlation of the genotypic and phenotypic values of the population (shown as average values over sequences of 10 generations).} \label{exp3-evol}
\centering
\end{figure}

To determine how competency might spontaneously evolve over generations, we introduced competency as an evolvable trait by letting each embryo’s competency level be determined by a single ‘competency gene’ with value in the range $[1, 500]$. During initialization, the competency genes of all embryos were set randomly to low values in the range $[1, 15]$. Then, during evolution cycles, we allowed each competency gene to be mutated, potentially taking values across the range of $[1, 500]$, and tracked the competency gene values of the best individual over 1000 generations (Figure \ref{exp3-evol}) The prevalence of the competency allele rapidly rose, meandering and exploring values up to 485 during evolution (shaded area in Figure \ref{exp3-evol}A) before plateauing at ~470. We provide a possible explanation for this outcome in the Discussions chapter.\\

To understand how allowing the competency gene to evolve over 1000 generations affects genotypic fitness, we looked at the phenotypic and genotypic values for the fittest individual in each generation (Figure \ref{exp3-evol}B). Values for the fittest individual quickly settled at consistent configurations in which the phenotypic and genotypic fitnesses diverged considerably. This is a fascinating outcome because it suggests that a certain level of competency reduces the pressure for improvements in an embryo’s structural genes. Once selection can no longer distinguish whether fitness is achieved by a set of good structural genes or by a high competency level that compensates for a poor set of structural genes, it can only improve the population by increasing competency, not by selecting better genetics.\\

\begin{figure}[htbp]
\includegraphics[width=7cm]{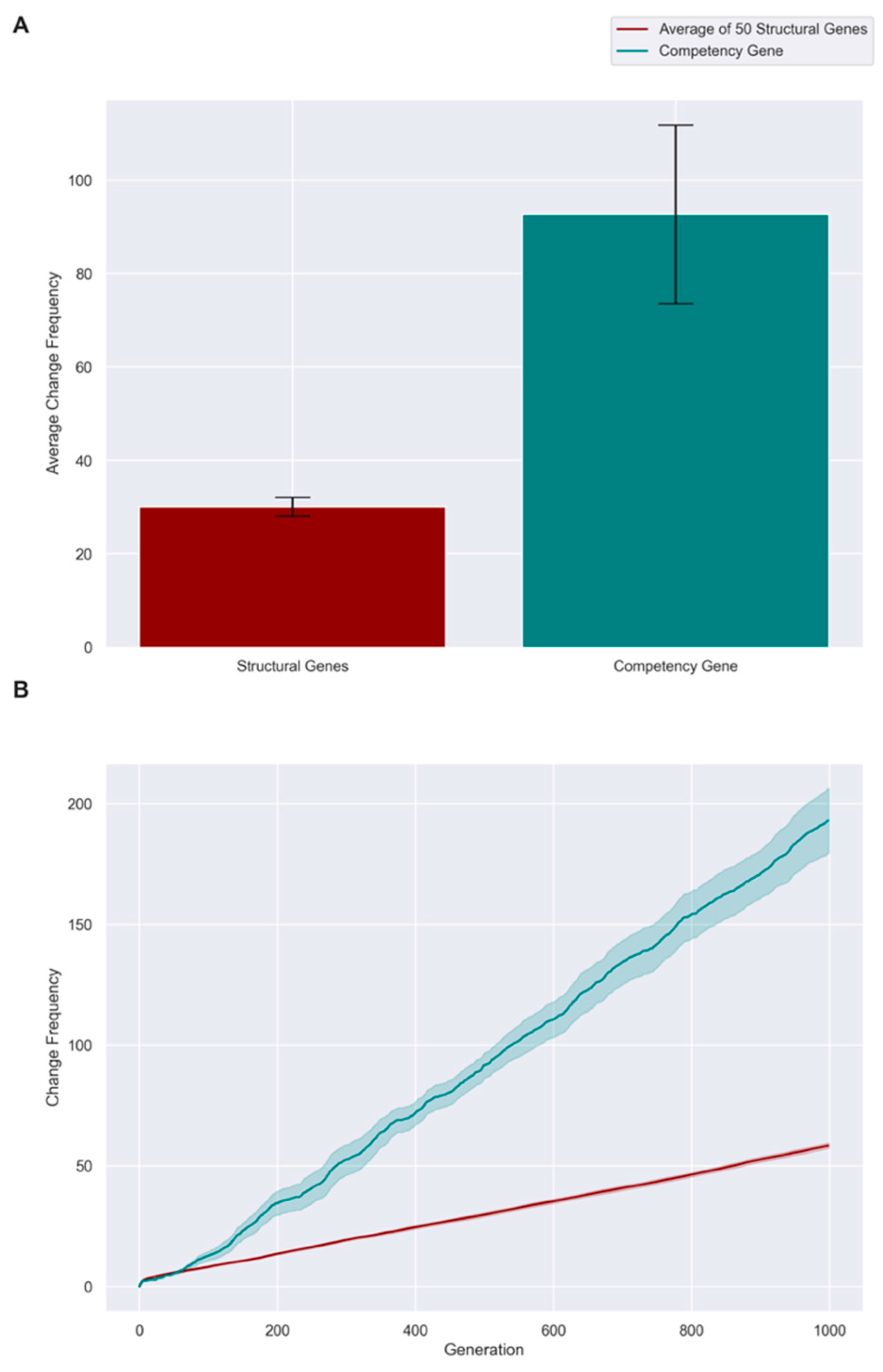} 
\centering
\caption[Gene Modification Frequencies in Experiment 3]{Evolution spends a greater proportion of time tweaking the competency gene compared to any structural gene. Figure recreated from \parencite{Shreesha2023}. Employing the experimental setup of Figure \ref{exp3-evol}, we checked how often changes occur within the structural genome of embryos vs. the competency gene, to determine where the evolutionary process focuses most of its effort under various conditions. \textbf{(A)}: Frequency of changes that 50 structural genes undergo versus the frequency of change that the single competency gene underwent, averaged over time. Error bars represent standard deviation over $n = 100$ repeat runs of the experiment. \textbf{(B)}: Comparison of frequency of changes in 50 structural genes versus the single competency gene, as a function of evolutionary time. The graph is cumulative, i.e., the number of changes made in the previous generation carry forward to the next. Shaded area represents variance over $n = 100$ repeat runs of the experiment.} \label{exp3-freqplot}
\centering
\end{figure}

To quantify this effect and determine how well selection, which ‘sees’ phenotypic fitness only, selects for genotypes when competency is allowed to evolve, we plotted the degree of correlation between genotypic and phenotypic fitness for all individuals in these populations (Figure \ref{exp3-evol}C). Correlation dropped to 0 within about 20 generations as individuals who succeeded because of their developmental competencies rapidly dominated the population. We conclude that allowing competency to evolve disrupts the ability to select for the best structural genes. We further validated this by examining the frequency, over 1000 cycles of evolution, with which positional changes to a single ‘cell’ resulted from tweaks to the competency gene vs. from tweaks to one of the structural genes. Figure \ref{exp3-freqplot} shows that the frequency of changes to the competency gene was much higher than the average of all fifty structural genes across 1000 generations in our simulation.

\subsection{Costs to Competency Ensures Genetic Assimilation: The Baldwin Effect}

The Baldwin Effect, as previously discussed, is the now broadly accepted fact in which individual organisms can achieve greater reproductive success based on behavioral adaptations, and that these adaptations can eventually become hardwired into the genome in subsequent evolutionary cycles.\\

Our initial simulations of the evolutionary impact of cellular competency did not exhibit the Baldwin Effect. This could have been due to the fact that our minimal model did not simulate any cost associated with increasing cellular competency, and thus there was no selective pressure towards genomic changes \parencite{Mayley1996, Mayley1996Conditions}. Although the actual energetic (or other) costs of cellular competencies are not known for any living model system, it is possible that the cellular computations required for axial patterning require additional resources over and above developmental events (competent or not) that are essential for any embryo. Thus, we next studied the effects of introducing a competency cost by penalizing the fitness of embryos in our model by a factor of their competency-value. Using penalty factors in the range of $[1 \times 10^{-7}, 0.5]$, we did see a Baldwin effect: the rate of rise of genotypic fitness corresponded positively with the increase in penalty factors. For penalty factors over 0.5, the genotypic fitness rose well above the phenotypic fitness, leading to disappearance of the Baldwin effect (see Appendix \ref{AppendixExpDetails} for a detailed overview)\\

\begin{figure}[htbp]
\includegraphics[width=12cm]{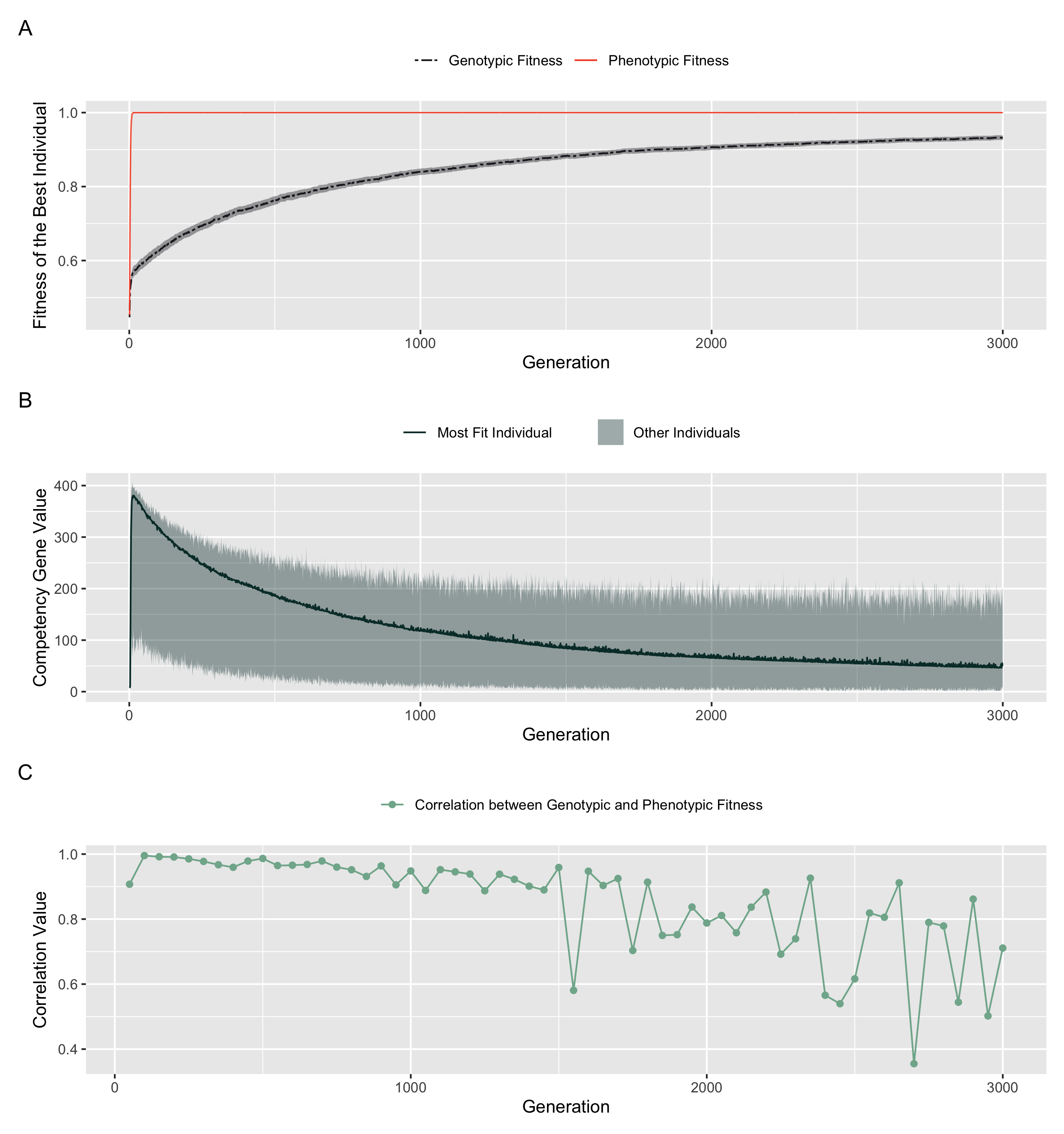} 
\centering
\caption[Experiment 4: Evolvable Competency with Cost]{Penalizing competency leads to its redundancy over time: the Baldwin Effect. Figure recreated from \parencite{Shreesha2023}. Competent embryos ($n = 100$) were initialized with an evolvable competency gene. At each developmental cycle, a fitness penalty of $1 \times 10^{-4}$ times the competency value was applied. \textbf{(A)}: Fitnesses of the best individual in a population of competent embryos with evolvable competency, penalized by a factor of $1 \times 10^{-4}$. Shaded area represents variance over 100 runs. \textbf{(B)}: Competency gene value chosen by the most fit embryo over the course of evolution. Shaded area represents the range of competency gene values in the population at each time point. \textbf{(C)}: Correlation of the genotypic and phenotypic values of the population (shown as average values over sequences of 50 generations).} \label{exp4-evolCost}
\centering
\end{figure}

The results of simulation using a penalty factor of $1 \times 10^{-4}$ over 3000 generations are shown in Figure \ref{exp4-evolCost}. As described above for simulations with no competency cost, phenotypic fitness reached its maximum in under 20 generations. However, unlike the previous experiment, the fitness of the structural genes did not plateau after a brief increase, but continued to improve over the course of evolution (Figure \ref{exp4-evolCost}A). Further, as the genotypic fitness rose, selection preferred progressively lower competency values (Figure \ref{exp4-evolCost}B). Phenotypic fitness was maintained at the maximum level, but embryos evolved to value structural genes over the competency gene. Over time, selection ensured that the genotype improved to a stage where competency became redundant -- the Baldwin effect. \\

We conclude that in the context of expensive competencies, selection is faced with a tradeoff between competency and the structural genome: it can either pick high competencies and bear subsequent penalties, or, it can pick low competencies and improve its structural genome. Since improving the structural genome does not bear a cost, selection prioritizes improvement of the structural genome, and over time, nullifies the effect of competency. Thus early gains based on the competency gene are later assimilated into the structural genes, paralleling what has been described previously in the context of organism-level learning \parencite{Mayley1996, Mayley1996Conditions, Turney2002}.
 
\chapter{Discussion} % Main chapter title

\label{Chapter5} % Change X to a consecutive number; for referencing this chapter elsewhere, use \ref{ChapterX}

% What did you do?

In this body of work, we evaluated the impact of morphogenesis on the rate and course of evolution by simulating the evolution of artificial embryos in-silico. We sought a minimalist model which captured a minimum of the morphogenetic competency as observed in real biology. We took particular inspiration from an experiment in-vitro, where scrambled organs of a tadpole, rearrange themselves into their normal positions over the course of development \parencite{vandenberg2012normalized}. To this end we allowed an array data structure, initialized with random cell values to sort itself in monotonic ascending order using a restricted form of bubble sort (see Chapter \ref{Chapter3} and Algorithm \ref{Alg2: development} for details) over the course of a "development" process. We included this developmental process as an additional step in a standard genetic algorithm (see Algorithm \ref{Alg1: GeneticAlg}) and observed its influence on evolution by monitoring fitness curves over time (see Chapter \ref{Chapter4} and Appendix \ref{AppendixExpDetails} for details of our experiments)

\section{Learning vs. Morphogenesis}

Our work is similar to those assessing the impact of learning on evolution. What is similar is the presence of a non-linear process (either learning or morphognesis) preventing the phenotype from being explicitly defined by the genotype. Indeed, such is the case in Biology as well: DNA specifies proteins, it exerts no direct control over morphology or function. What is crucially different is the scale at which these mechanisms function. Learning is an animal-scale trait, requiring specific structures such as wired nervous systems to facilitate learning. Our model makes no such demands, arising from the basic property of cells to sense their neighbors and move to positions of least stress \parencite{levin2022collective}.\\

However, morphogenetic-competency examined here and behavioral learning, explored by others can interact with each other. For instance, cellular collectives could be influenced in their signaling by the manner in which an organism interacts with its environment (behavior). Environmental signals could seep into lower levels and shape competency towards specific functional outcomes; this could be a potential explanation for how phenotypic adaptive plasticity manifests.

\section{Impact of Competency on Evolution}

We found that providing cells with minimal competency to improve their position in the virtual embryo results in better performance of the evolutionary search. Populations reach better fitness values faster when cellular activity is able to make up for genetic deficiencies (Figure \ref{exp1:pheno}). Indeed, in mixed populations, competent individuals tend to dominate and rapidly take over (Figure \ref{exp2:mixed}) as long as they have a minimal level of competency and/or are present in adequate numbers (Table \ref{Table2: mixed domination}). The simulation highlighted the distinction between two properties of each individual that are often conflated or obscured in simulations that do not include an explicit competency step: genotypic vs. phenotypic fitness.\\

Perhaps the most interesting aspect was the role that competency played in exacerbating the inability of selection to evaluate the genetic material that gets passed on to subsequent generations. We observed that increases in competency made it harder and harder for selection to pick the best structural genes. Specifically, the correlation between genotypic and phenotypic fitness drops to insignificant levels very rapidly (Figure \ref{exp3-evol}C). This could be expected to result in complex dynamics, because competency improves fitness of individuals but impairs the ability of the evolutionary hill-climbing search in fitness space to pick out the most elite structural genomes. Thus, we studied what happens when evolution is also allowed to control the degree of competency, which is biologically realistic since cellular capacities for sensing, computation, and action are themselves under evolutionary selection. We observed that the population drives towards picking the highest competency gene value in the population (Figure \ref{exp3-evol}A). This suggests evolution's tendency to rely on competency rather than raw structural genomes. Evolution simply does not have to improve if a "bad quality genome" encodes high competency. We elaborate on this in the next section.\\

In our models, we had to make a number of quantitative choices with respect to the evolutionary process. Thus, we checked how sensitive our conclusions were to these decisions via a hyperparameter scan: re-running the simulations with different choices for various hyperparameters (see Appendix \ref{AppendixExpDetails}). Specifically, we identified mutation probability and selection stringency as key hyperparameters which could influence the results of evolution. In an effort to probe their influence on the final competency gene value attained, we ran this experiment for 132 different combinations of mutation probability and selection stringency in the range of $[0.2, 0.8]$ and recorded the stable-competency value attained for each hyperparameter combination (Figure \ref{fig:Appen-hyperparameterTest} in Appendix \ref{hyperparametersection-appendix}). Correlation analysis revealed that a correlation of $-0.4$ existed between mutation probability and stable-competency-gene-value. However, no relationship was found between selection stringency and the stable-competency-gene-value. A possible reason for this could be that after generation 20, almost every embryo in the population achieves maximum phenotypic fitness, therefore there is no difference in choosing the top 20\% of the population or the top 80\% of the population. Mutation probability on the other hand has a direct influence on changing individual fitness, which explains its moderately significant relationship with the stable-competency-gene-value.

\section{Evolution: An Intelligence Ratchet}

Evolution ignores genome quality when given the option to boost intelligence i.e., morphological competency. Manually raising competency led to progressive worsening of genomes (Figure \ref{exp1:geno}). This was further confirmed when evolution was allowed to choose competency (Figure \ref{exp3-evol}). It would seem that evolution gets progressively locked in to improving the agential material with which it works, with reduced pressure on the structural genes, ratcheting up effort into the developmental software -- emergent communication networks between cells -- than perfecting the hardware (cellular machinery). Planaria exemplify such behavior: wild planarian, because of reproduction by fissioning and regeneration have a chaotic genome, with different chromosomes in each cell; yet they have the most reliable anatomy -- regenerating to a complete worm each time. It would seem that planarian genomes rely on the competency of their cells, rather than perfecting their structural arrangement to exhibit such reliable behavior.\\

In order to understand where evolution was applying its "effort", we checked to see how often the structural genes were tweaked vs. how often the competency gene was tweaked during the course of evolution (Figure \ref{exp3-freqplot}). We found that the competency gene changed significantly more often than any other structural gene. This further validates the presence of an intelligence ratchet. \\

Given that morphogenesis is capable of coordinating intelligence in multiple sub-spaces (transcriptional, cellular etc.) and at multiple-scales -- molecular, cell, tissue, organ \parencite{fields2020morphological}. The intelligence ratchet could be a driver of scaling intelligence. In our simulations, intelligence was considered only in the morphological domain, but at two scales: at the cellular level -- where each cell served to rearrange itself locally -- and one at the organism level, where a fitness function ensured selection of arrays with the most order. By ratcheting up competency, evolution encouraged local re-organization, giving rise to global order. A similar principle could be occurring in multiple-subspaces and at multiple-scales. We seek to assess this claim in future work.

\section{Costs of Competency}

One factor, which we had not considered in our initial framework was the costs of competencies. Despite this omission, we found that evolutionary dynamics were significantly impacted by competency. It would seem then that the Baldwin effect \parencite{Baldwin1896} and genetic assimilation \parencite{Waddington1953} are not the only possible ways in which competency can influence evolution. \\

Costs, when considered from the perspective of learning clearly apply: it takes time, effort, and resources to gain knowledge of the environment; given this cost, evolution is incentivized to genetically assimilate those traits which help with learning (the Baldwin effect). However, costs of morphogenesis -- which we consider as competency here -- aren’t as clear. Whether morphogenesis bears a cost, remains to be determined by measurements in vivo that have not yet been done. It could be reasonable to claim that morphogenesis bears metabolic or other costs, but it is equally possible that the competencies of cells bear no penalty, existing by default -- a process of self-organization which utilizes internal pre-existing processes. \\

Nonetheless, we did include a cost to our framework and noticed that the Baldwin effect does indeed manifest (Figure \ref{exp4-evolCost}), confirming earlier reports that the Baldwin effect requisites a fitness-penalty \parencite{Mayley1996, Mayley1996Conditions, Turney2002}. However, we remain skeptical of this result. While the biological relevance of this simulation is pertinent to behavioral competency, it might not be so for morphogenesis. We hope future work assessing the costs of morphogenesis will help clarify this issue.

\section{Morphogenesis in Evolutionary Algorithms}

Our results indicate that regulative morphogenesis (the ability of cellular behaviors to adjust phenotype toward a specific outcome, despite their genetically-determined initial states) can boost evolutionary capacities towards better structural outcomes. Given the intimate relationship between form and function, it is likely that morphogenesis serves as a director of functional outcomes in Biology. However, efforts in Machine Learning, Evolutionary Robotics, and Artificial life, have failed to leverage its capabilities. Works such as NEAT \parencite{stanley:ec02} and its variants \parencite{HyperNeat} address the form-function relationship in part, but consider the genotype a direct map of the phenotype, individual neurons as passive entities, and intelligence as an observable trait in the 3D world. If we are to capture intelligence as exhibited by Biology, it serves to imitate biological processes heavily, accounting for the emergence of problem-solving ability rather than handcrafting it through high level structures such as neural networks. \\

To this end, we propose a new evolutionary framework for \textit{growing} collective 
intelligent systems using evolutionary algorithms. Genotypes within such a system would serve to construct goal-directed agential elements allowing their resulting coordination at multiple-scales to self-assemble into a morphological structure capable of robust functionality. Evolution would be tasked with choosing those genomes which develop best for a pre-defined task.\\

A key aspect within such a system would be the external environment. In our present work, we did not consider one -- our goal was to assess the role of morphogenesis on evolution without interference from the environment. However, if one were to grow collective intelligences, a complex environment whose dynamics mirrors our own, at least to the extent of allowing agents to solve a task through multiple means, must be considered. 

\section{Limitations and Future Work}

Our framework was more complete than many evolutionary simulations because it included an explicit developmental layer between the genotype and phenotype. It was multiscale in the sense that important changes occurred on an evolutionary scale across individuals, but also ones driven by components of those individuals within their lifetime -- the cells, which had their own perspective and local goals. However, our system clearly omitted a huge amount of biological detail with respect to cellular mechanisms of sensing, competition, cooperation, etc. We intentionally designed a minimal model to specifically focus on a few sufficient dynamics, and this likely under-emphasized the difference between cellular competencies and, for example, effects of learning at the organism level on evolution. Fundamentally we explored a toy model virtual world in which the individual roles of selection and competency could be quantitatively dissected in the absence of confounding complexity. We sought generic laws and dynamics, not a simulation of the detailed trajectory of any existing biological species.\\

Future work will add physiological layers, diverse cell types, computation at gene-regulatory and cellular-network levels, and a multi-dimensional target morphology (e.g., 2D or 3D pattern instead of just one primary axis) to more closely model biological reality. There is also much that can be improved with respect to the specific mechanisms that cells use to implement their competency: a rich set of diverse genes will be added in the future to enable evolution to manipulate different types of local goals and competencies.
 
\chapter{Conclusion} % Main chapter title

\label{Chapter6} % Change X to a consecutive number; for referencing this chapter elsewhere, use \ref{ChapterX}

Our results suggest an interesting interplay between morphogenetic-competency and evolution other than the one conventionally studied by Baldwin \parencite{Baldwin1896} and the rest of the evolutionary computation community \parencite{Hinton1996, Belew1990, Gruau1993, Nolfi1999, Nolfi1994LearningAE, Whitley1994, Mayley1996, Turney2002, Bull1999, French1994, CarseO00, Parisi19911L, Ku2006}. 

It hints at a host of factors (in addition to behavior and morphogenesis) which could be manipulating evolutionary dynamics, driving its efficiency to one capable of crafting superior problem-solving machines. Intelligence (problem solving competency) could have been an evolutionary driver long before complex brains and muscle-driven behavior arose \parencite{keijzer2013nervous, keijzer2015moving, Lyon2021}. Beyond understanding natural evolution, acknowledging the multi-scale problem solving capability of living systems from the perspective of morphogenesis \parencite{Levin2023} could help serve the computer science community at large in its quest for generalizable intelligence. 
%----------------------------------------------------------------------------------------
%	THESIS CONTENT - APPENDICES
%----------------------------------------------------------------------------------------

\appendix % Cue to tell LaTeX that the following "chapters" are Appendices

% Include the appendices of the thesis as separate files from the Appendices folder
% Uncomment the lines as you write the Appendices

% Appendix A

% Appendix Template

\chapter{Software and Hardware Setup} % Main appendix title

\label{AppendixSoftware} % Change X to a consecutive letter; for referencing this appendix elsewhere, use \ref{AppendixX}

This chapter lists the hardware, software, and software-libraries used in our simulations.

\section{Software}

Our codebase is written in python. It was a natural choice given the exploratory nature of this project. We provide a thoroughly documented open-source repository for the manipulation of arrays for morphogenetic simulation. Our code can be found here: \href{https://github.com/Niwhskal/CellularCompetency}{https://github.com/Niwhskal/CellularCompetency}

\subsection{Python Packages}

Numpy arrays (float32) were used to carry out array manipulation for morphogenesis. Matrices of such arrays served as populations for our genetic algorithm (algorithm \ref{Alg1: GeneticAlg}). \\

Fitness values (genotypic and phenotypic), from each generation were written to .npy files, and were subsequently plot using the Matplotlib and Seaborn libraries. \\

Reproducibility was ensured with the Numpy random-number-generator, initialized with a seed.

\section{Hardware}

Experiments were carried out on the M1 Macbook Pro (2020, 8GB).

% Appendix Template

\chapter{Experimental Details} % Main appendix title

\label{AppendixExpDetails} % Change X to a consecutive letter; for referencing this appendix elsewhere, use \ref{AppendixX}

\section{Hyperparameters}

The following served as main hyperparameters in our simulation:
\begin{itemize}
    \item[1.] Size of each Embryo: Chosen to be 50. Simulations revealed no significant impact on evolution post an embryo size of 45.

    \item[2.] Population size: Chosen to be 100 mainly to enforce a cap on compute. However, simulations run with a population size of 10,000 revealed no significant impact on evolution.

    \item[3.] Number of GA cycles: unless otherwise specified, it was set at either 3000 or whenever an embryo reached maximum phenotypic fitness, whichever was earliest.

    \item[4.] Mutation probability of genes: was chosen to be 0.6 based on a hyperparameter scan (see section \ref{hyperparametersection-appendix} below for an overview).

    \item[5.] Selection Stringency: was chosen to be 0.1 based on a hyperparameter scan. (see section \ref{hyperparametersection-appendix} below for an overview)
\end{itemize}

\section{Simulation Details}

\subsection{Experiment 1: Evolving a single hardwired population with three different competent populations}

\subsubsection{Common specifics of hardwired or competent population}

\begin{itemize}
\item arraySize = 50
\item cell value initialization range = [1, 50]
\item selection stringency = 10\%
\item number of embryos in the population = 100
\item max. Generation = 1000
\item mutation Probability = 0.6
\item number of repetitions = 100
\end{itemize}

\subsubsection{Specifics of competent population}
\begin{itemize}
    \item Competency value = 20 or 100 or 400 swaps
\end{itemize}

Phenotypic and genotypic fitnessess of the best “individual” in each of the four populations (one hardwired, three competent) in each generation were plot over 250 generations (Figures \ref{exp1:pheno} and \ref{exp1:geno} respectively in the paper).\\

Shaded areas indicate 95\% confidence interval bands over 100 repeats of the experiment.

\subsubsection{Statistical Significance}

Based on the results of Experiment-1 (Figure \ref{exp1:pheno}, in the paper), we deduced that increasing the competency value led to a higher rate of fitness increase compared to lower competency or even zero competency (hardwired) values.
We verified this observation statistically by employing a t-test: Experiment-1 was repeated 100 times and fitness curves were compared at several generations. We compared the hardwired (genotypic) fitness to each of the competent (phenotypic) fitnesses at generations 2, 10, and 20. These specific generations were chosen because the variance in repeats around them was the greatest. At each of these generations, the conditions to carry out a t-test were verified: it was ensured that our repeats were gaussian distributed, and that they shared similar variances (the variance ratio between any two distributions was ensured to be less than 1:4). We used different random seeds for each experimental run to ensure that they were independent of each other. \\

Results from our t-tests are recorded in Table \ref{pvalue-Table} (we notice the same p-values for generations 2, 10, and 20)

\begin{center}
\begin{table}
\caption[T-test results: Experiment 1]{Results of a t-test performed to assess the difference between hardwired and competent populations. P-values in the same order of magnitude were noticed for each of the three generations. Table recreated from \cite{Shreesha2023}} \label{pvalue-Table}
\begin{tabular}{|c|c|} 
 \hline
 Phenotypic fitness [Competency 20] vs. Hardwired & p-value << 0.001 \\ 
 \hline
 Phenotypic Fitness [ Competency 100] vs. Hardwired  & p-value << 0.001\\  
 \hline
 Phenotypic Fitness [ Competency 400] vs. Hardwired  & p-value << 0.001 \\ 
 \hline
\end{tabular}
\end{table}
\end{center}

\subsection{Experiment 2: Evolving multiple hybrid populations, each composed of hardwired and competent embryos}

\subsubsection{Common specifics of a hybrid population}
\begin{itemize}
    \item arraySize = 50
    \item cell value initialization range = [1, 50]
    \item max. Generation = 30
    \item selection stringency = 10\%
    \item total number of embryos (regardless of kind) = 200
    \item mutation probability = 0.6
    \item number of repetitions = 20
\end{itemize}

At the start of evolution, hardwired and competent embryos were combined together into a single mixed population (Figure \ref{exp2:mixed}). Each embryo developed into an individual according to its respective developmental cycle. \\

Selection occured in a combined fashion: phenotypic fitnesses (same as the genotypic fitness) of hardwired individuals were compared against phenotypic fitnesses of competent individuals to determine the fittest 10\% of the hybrid population.\\

During the process of cross-over, selected hardwired and competent embryos reproduced to repopulate the hybrid population to its original strength (200 in our experiments). Hardwired and competent embryos do not interbreed. The repopulated population was subject to point mutations at random locations, post which a next evolutionary cycle began.\\

Prevalence (percentage) of hardwired and competent embryos in each generation were plot over 30 generations (Figure \ref{exp2:mixed}). Shaded areas around each curve represents the variance over 20 experimental repeats.

\subsection{Experiment 3: Evolving a competent population with evolvable competency} \label{hyperparametersection-appendix}

\subsubsection{Competent population specifics}
\begin{itemize}
    \item arraySize = 50
    \item cell value initialization range = [1, 50]
    \item selection stringency = 10\%
    \item number of embryos in the population = 100
    \item max. Generation = 1000
    \item mutation Probability = 0.6
    \item number of repetitions = 100
    \item competency = \textit{evolvable}
\end{itemize}

The competency value of a competent embryo was set to be evolvable. i.e, each array carried an extra cell at position n+1 whose value was set to indicate its competency value. \\

At generation 0, competent embryo’s were initialized with competency values in the range of [1-15]. Its function was solely to provide this value; it was not considered part of the “morphological structure” of the array. Post generation 1, we allowed this competency cell to be mutated to a value in the range of [1-500] with a probability of 0.6.\\

Fitness's of the best individual in the population (Figure \ref{exp3-evol}B), together with its respective competency value (called the competency gene value in the paper) (Figure \ref{exp3-evol}A) were plot over 1000 generations. The shaded area in Figure \ref{exp3-evol}A indicates the range of competency gene values which were prevalent at any generation. \\

\begin{figure}[h]
\includegraphics[width=12cm]{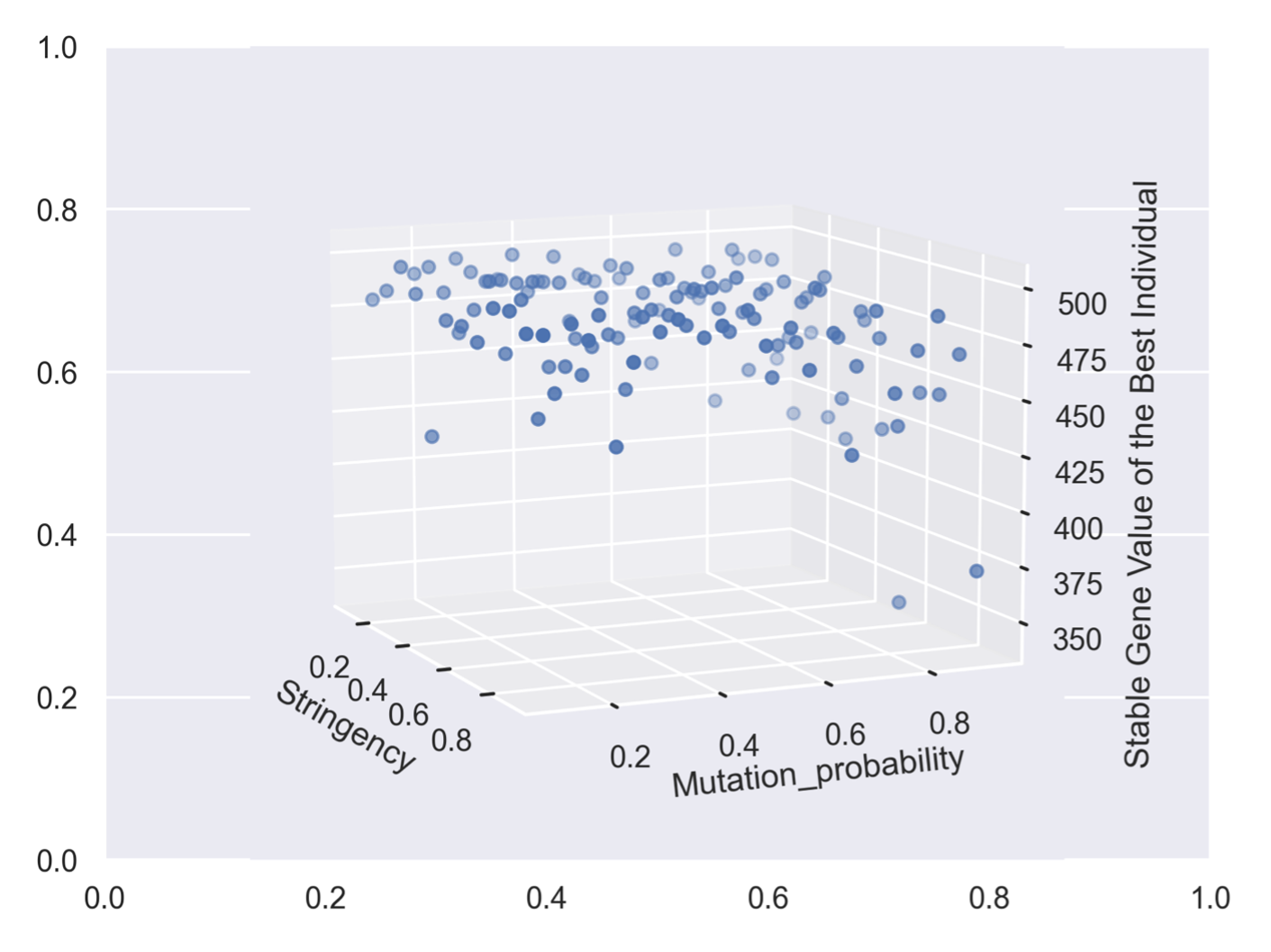} 
\centering
\caption[Mutation Probability and Selection Stringency Hyperparameter Scan]{Results of varying mutation probability and selection stringency on the final stable competency gene value. 132 different combinations of mutation probability and selection stringency in the ranges of [0.2, 0.8] were chosen for each experimental run. Each point represents the stable competency gene value attained for a specific combination of mutation probability and selection stringency. Figure recreated from \cite{Shreesha2023}} \label{fig:Appen-hyperparameterTest}
\centering
\end{figure}

Further, we assessed the role of hyperparameters on the value at which the competency gene stabilizes. Specifically, we checked the effect of mutation probability and selection stringency on the final stable competency gene value attained (Figure \ref{fig:Appen-hyperparameterTest}).

\begin{figure}[htbp]
\includegraphics[width=12cm]{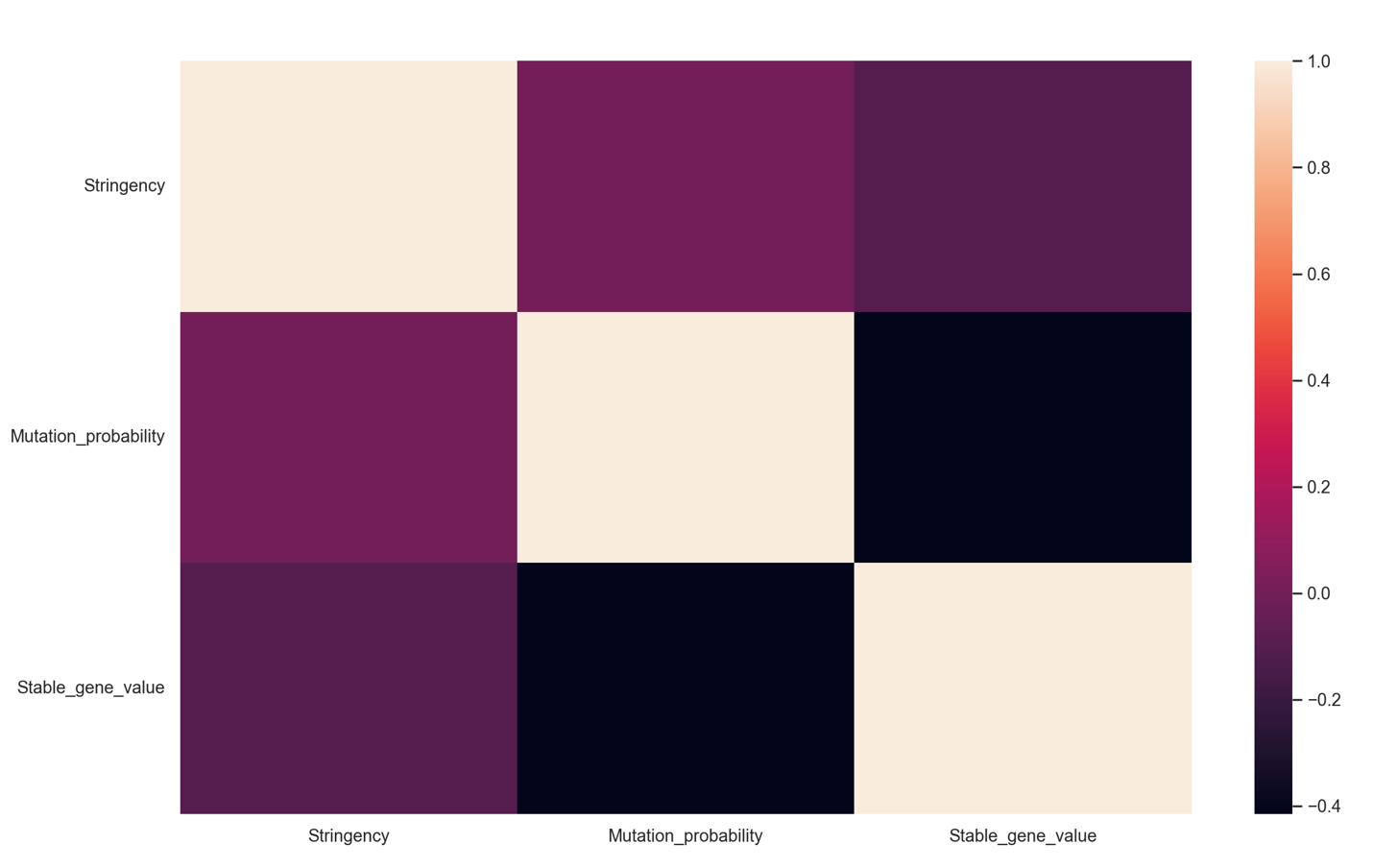} 
\centering
\caption[Correlation Matrix of Hyperparameters]{Correlation matrix of mutation probability, selection stringency, and stable-competency-gene value attained.} \label{fig:Appen-correlation}
\centering
\end{figure}

Visually, we did not notice any distinct relationships between these hyperparameters and the stable-competency-gene-value attained. We carried out a correlation analysis of each of these variables to further probe the issue (Figure \ref{fig:Appen-correlation}).\\

The correlation matrix indicated no correlation between selection stringency and stable-competency-gene-value attained. However, a minor negative correlation (-0.4) was found between mutation probability and the stable-competency-gene-value attained. \\

Finally, we checked to see how often the “structural cells” of an array (the part which is assessed for its order) get modified versus how often the competency cell / gene gets modified over the course of evolution. \\

At each generation, each of the 50 structural cells of the array were checked to see how often they get modified. At the end of 1000 generations, we had a count of how often each of the 50 structural genes changed, and how often the competency cell / gene of the array changed. \\

We took the average of these 50 structural-cell counts, and compared them to the competency-gene-change counts at each generation (Figure \ref{exp3-freqplot}B in the paper), and at the end of 1000 generations (Figure \ref{exp3-freqplot}A in the paper).\\

\subsection{Experiment 4: Including Costs to competency}

\subsubsection{Competent population specifics}
\begin{itemize}
    \item arraySize = 50
    \item cell value initialization range = [1, 50]
    \item selection stringency = 10%
    \item number of embryos in the population = 100
    \item max. Generation = 3000
    \item mutation Probability = 0.6
    \item number of repetitions = 100
    \item competency = evolvable
    \item competency penalty weight ($pw$) = 1e-04
    \item max competency gene value allowed ($x_{max}$) = 500
\end{itemize}

Here, as before, we set the competency value to be evolvable but each competency value carried a cost. Given a competent embryo, $C$, carrying a competency value of $x$, we first normalize the competency as follows:

\begin{equation}
    c^{'} = \frac{x}{x_{max}}
\end{equation}

where, $x_{max}$ is the maximum competency value allowed. $C$'s phenotypic fitness (pf) is penalized as:

\begin{equation}
    pf^{'} = pf - pw * c^{'} 
\end{equation}

Figure \ref{exp4-evolCost} revealed that evolution preferred to reduce its dependence on competency because it bore a cost, choosing instead to improve the structural genes to improve morphology -- a clear indication of the Baldwin effect.

%----------------------------------------------------------------------------------------
%	BIBLIOGRAPHY
%----------------------------------------------------------------------------------------

\printbibliography[heading=bibintoc]

%----------------------------------------------------------------------------------------

\end{document}